\documentclass[sigconf,nonacm,balance=false]{acmart}

\AtBeginDocument{%
  }

\begin{document}

\title{OrganLens: Organ-Specific Representation Learning for CT Foundation Models}

\author{Zhixuan Ge}
\authornote{These authors contributed equally to this work.}
\email{zg33@rice.edu}

\author{Anqi Li}
\authornotemark[1]
\email{al318@rice.edu}
\affiliation{%
  \institution{Rice University}
  \city{Houston}
  \state{Texas}
  \country{USA}}

\author{Sadeer Al-Kindi}
\affiliation{%
  \department{Department of Cardiology}
  \institution{DeBakey Heart and Vascular Center}
  \city{Houston}
  \state{Texas}
  \country{USA}}
\email{sal-kindi@houstonmethodist.org}

\author{Hanwen Xu}
\authornote{Corresponding author.}
\affiliation{%
  \institution{University of Pennsylvania}
  \city{Philadelphia}
  \state{Pennsylvania}
  \country{USA}}
\email{xuhw@engineering.upenn.edu}

\author{Wei Qiu}
\authornotemark[2]
\affiliation{%
  \institution{Rice University}
  \city{Houston}
  \state{Texas}
  \country{USA}}
\email{wq8@rice.edu}

\renewcommand{\shortauthors}{Ge et al.}

\begin{abstract}
A CT examination captures multiple organs, but many biomedical questions
concern abnormalities, prognosis, or longitudinal change in a specific organ.
These questions require a separate representation for each organ within the
same CT volume. Existing CT foundation models commonly produce a single
volume-level representation, while recent anatomy-aware methods either
encode pre-separated organ volumes or explicitly disentangle images into
organ token groups. The former may remove clinically relevant surrounding
context, while the latter does not condition a shared encoder on a selected
organ before its features are formed. We introduce OrganLens for
organ-specific representation learning through self-supervision. An organ
identity conditions a shared CT encoder, while organ-specific
distillation and anatomy-mask supervision shape features for
anatomy-weighted pooling into organ-specific representations. At inference, the shared model
produces 11 organ-specific representations without external segmentation masks.
We evaluate OrganLens on CT-RATE, RAD-ChestCT, INSPECT, and NLST across
diverse acquisitions and downstream evaluations. Relative to
the strongest baseline, heart representations raise CT-RATE cardiomegaly AUROC
from 0.917 to 0.953, while lung representations improve the Harrell C-index
for NLST lung-cancer, respiratory, and COPD mortality by 0.003, 0.035, and
0.030. The global representation reaches
INSPECT Recall@10 of 33.09\% and 32.04\% for text-to-image and image-to-text
retrieval, respectively. Across organ-related tasks, anatomically matched
representations provide stronger task-relevant signal, while the global
representation retains broad utility. OrganLens offers a scalable approach to
organ-specific CT representation learning with a shared encoder. More broadly,
it provides the medical research community with a reusable framework for
studying organ-specific disease across cohorts and clinical endpoints. Code and pretrained models are available at
\url{https://github.com/gezhixuan/OrganLens}.
\end{abstract}



\maketitle

\section{Introduction}
CT captures multiple organ systems and their spatial relationships in a single
examination, supporting analyses of abnormalities, prognosis, and longitudinal
change that often target a specific organ. Such analyses call for organ-specific representations, as disease processes and clinically relevant phenotypes may vary substantially across organs within the same individual. Recent CT foundation models learn transferable
representations across visual and vision--language tasks
\cite{hamamci2024developing,blankemeier2026merlin,
yang2025crcfound}. However, their
general-purpose representations are typically defined at the volume level
rather than conditioned on a selected organ. 
Because a volume-level representation must integrate organs that differ in size, appearance, and relevance to a given endpoint, it dilutes organ-specific evidence.

Existing anatomy-aware approaches obtain organ specificity in three ways.
Encoding pre-segmented organ inputs isolates anatomical structures
\cite{yamamoto2026totalfm}, but may discard surrounding context relevant to
abnormality detection. Disentangling images into organ-wise token groups
recovers organ-specific components \cite{song2025owt}, but without explicit
conditioning on a selected organ, the learned groups may retain irrelevant
features and dilute organ-specific signal. Localizing anatomy through
text-derived supervision \cite{lin2024ctglip,shui2025large,you2026oka} depends
on report--region alignment, which may lack the spatial precision needed to
distinguish adjacent organs. A separate encoder for each organ would also scale
poorly and prevent representation sharing.
The goal is therefore not to locate or isolate an organ, but to learn a shared encoder that can be conditioned on a specific organ, adaptively select which features belong to it, and produce organ-specific representations. This raises a central question:
can one CT encoder adapt its representation to a selected organ while
preserving surrounding anatomical context and producing anatomy-weighted
representations without manual segmentation at inference?

To answer this question, we introduce OrganLens, an organ-specific
representation learning framework built on a shared CT-pretrained
DINOv2 encoder~\cite{oquab2023dinov2,xu2025cardiac}. OrganLens combines three components. \textit{Organ-identity conditioning}
adapts encoder features to each of 11 organs.
\textit{Organ-guided cropping} retains the selected anatomy in
self-distillation views. \textit{Anatomy-mask supervision} trains a spatial
decoder to predict an organ-specific anatomical mask. At inference, this
mask provides soft weights for pooling patch features without external
segmentation masks. OrganLens also retains regions surrounding the target
organ, preserving useful anatomical context. The model produces
11 organ-specific representations for downstream analyses.
Here, organ-specific denotes organ-conditioned task preference.
We evaluate these representations on CT-RATE~\cite{hamamci2024developing},
RAD-ChestCT~\cite{draelos2021machine}, INSPECT~\cite{huang2023inspect}, and
the National Lung Screening Trial (NLST)~\cite{national2011national}. These
cohorts span unenhanced CT, CT pulmonary angiography, and low-dose screening
and support four task families: abnormality detection, prognostic prediction,
longitudinal analysis, and image--text retrieval.

Across these evaluations, OrganLens generally outperformed the competing
methods, including CT-pretrained DINOv2 and
GigaHeart~\cite{xu2025cardiac}. Anatomically matched representations obtained
the highest macro-AUROC on both abnormality detection cohorts, demonstrating
the benefit of organ-specific representation learning.
For prognosis, anatomically matched representations ranked first among
OrganLens representations for 16 of 17 endpoints, and OrganLens achieved the
highest C-index point estimate in 11 of 12 endpoint-visit comparisons in the
longitudinal analysis.
Together, these results indicate that predictive signal was concentrated in
anatomically matched organs rather than distributed uniformly across the entire volume.
Beyond organ-specific analyses, the global OrganLens representation performed
best in image--text retrieval on CT-RATE and INSPECT, suggesting that the combined organ-specific representations anchor image--text
correspondence to individual organs rather than coarse anatomical regions.
These findings suggest that OrganLens could provide a foundation for future
clinical applications involving organ-specific disease characterization, risk
stratification, and longitudinal monitoring.

Our contributions are: 
\begin{itemize}
    \item We introduce OrganLens, which conditions a shared CT encoder on organ identity, learns organ-specific features while retaining surrounding anatomical context, and produces 11 organ-specific representations.
    \item We develop a self-distillation framework that combines organ-identity
    conditioning, organ-guided cropping, and anatomy-mask supervision to produce
    organ-specific representations through anatomy-weighted pooling.
  \item We demonstrate the clinical value of learning multiple organ-specific representations from the same CT examination: anatomically matched representations provide finer-grained, task-relevant features for organ-focused downstream applications, while their global aggregation retains broad utility for whole-volume tasks.
\end{itemize}

\section{Related Work}
\begin{figure*}[!t]
  \centering
  \includegraphics[width=\textwidth]{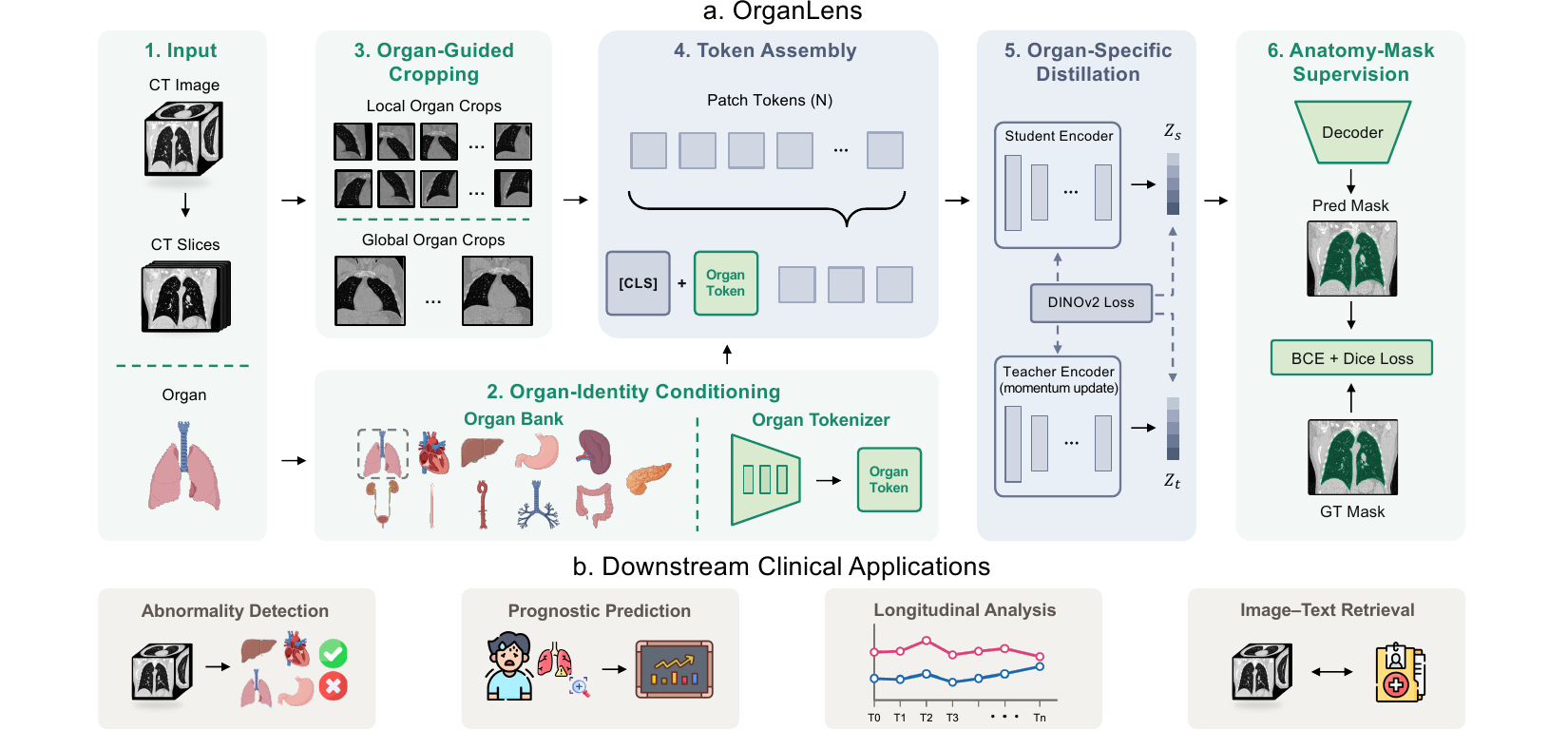}
    \caption{Overview of OrganLens. (a) Organ-identity conditioning, organ-guided
cropping, organ-specific distillation, and anatomy-mask supervision produce
anatomy-weighted organ-specific representations. (b) The representations are reused across
    downstream clinical applications.}
  \label{fig:organlens-overview}
\end{figure*}

\subsection{CT and Volumetric Foundation Models}
Foundation models provide reusable representations across endpoints
\cite{bommasani2021opportunities}. CT-specific self-supervision spans 3D image
restoration in Models Genesis, hierarchical pretext tasks in Swin UNETR,
unified 2D--3D training in UniMiSS, and subvolume context modeling in VoCo
\cite{zhou2021models,tang2022self,xie2022unimiss,wu2024voco,
yang2025crcfound}. DINO and DINOv2
learn transferable ViT features through self-distillation, masked patch
prediction, and representation regularization
\cite{caron2021emerging,zhou2022ibot,sablayrolles2019spreading,
oquab2023dinov2}.

CT-CLIP aligns CT-RATE scans and reports for zero-shot detection and retrieval
\cite{hamamci2024developing}. Merlin scales CT vision--language learning using
a large clinical resource \cite{blankemeier2026merlin}, while FM-HCT transfers
a self-supervised head CT encoder across detection tasks \cite{zhu20263d}.
SPECTRE combines DINO-style self-supervision with report-level
vision--language alignment in a hierarchical volumetric transformer
\cite{claessens2025spectre}.
SigVLP uses axial rotary position embeddings for variable-depth volumes and aligns
3D chunks with organ-wise report observations \cite{wang2026sigvlp}. These
systems provide local or volume-level features and include both general and
anatomy-specific models.

Recent CT models include staged anatomical and semantic pretraining in FlexiCT,
volumetric DINO self-distillation in CoralBay, and lung-focused pretraining in
UCLIF. LCTfound uses diffusion pretraining with imaging and clinical
information for lung CT \cite{li2026flexict,gatopoulos2026coralbay,
li2026uclif,gao2026lctfound}. TANGERINE uses computationally efficient 3D
masked-autoencoder pretraining for volumetric thoracic CT
\cite{mcconnell2026tangerine}. The CIPHER preprint adapts contrastive
masked-autoencoder pretraining to pretreatment pneumonitis-risk prediction
\cite{muneer2026cipher}. Across this literature, representations are not
indexed by multiple organ identities applied to the same examination.

\subsection{Anatomical Localization and Promptable Segmentation}
Anatomical segmentation provides explicit spatial priors for focused CT
analysis. TotalSegmentator delineates many organs and structures in CT;
MedSAM, SegVol, and ONCOPILOT provide promptable segmentation for medical
images, 3D volumes, and solid tumors, respectively
\cite{wasserthal2023totalsegmentator,ma2024segment,du2024segvol,
machado2025oncopilot}.
Language-driven universal segmentation further uses semantic organ and tumor
labels to parameterize a shared CT model \cite{liu2024universal}. For
end-to-end abdominal CT classification, ORACLE-CT uses multi-organ
segmentations to constrain label-specific attention pooling over encoder
features \cite{dahal2026oraclect}. These anatomical masks support organ
cropping, regional pooling, and localized supervision. These workflows leave
feature formation independent of the selected identity. Cropping narrows the
available context, post hoc pooling selects already-formed features, and
external inference masks introduce preprocessing and segmentation domain
shift. OrganLens uses masks during pretraining and predicts its own spatial
mask at inference.

\subsection{Anatomy-Aware and Multi-Organ Representation Learning}
Anatomical priors can also shape the representation itself. Adam-v2
\cite{taher2024representing} models part--whole relations through objectives
for localizability, composability, and decomposability. DrasCLR organizes lung
CT features by anatomical location \cite{yu2024drasclr}, GigaHeart specializes
features for heart-transplantation assessment \cite{xu2025cardiac}, and
BoneCoT learns across skeletal sites and bone-metastasis tasks
\cite{zhao2026bonecot}. These methods cover relational, location-aware, cardiac, and skeletal
representations with fixed anatomical scopes or representation schemes.

Beyond CT-specific models, Oh et al. condition ultrasound transformer features
on anatomical context, while AG-SSD combines anatomy-consistent cropping and
self-distillation for CT and MRI \cite{oh2026anatomy,yu2026agssd}. Both show
that anatomical context can guide feature formation and self-distillation
across imaging modalities.

Coda models variation and consistency across 23 radiographic regions
\cite{hosseinzadeh2025learning}. OWT disentangles CT or MRI into composable
organ-token groups, whereas TotalFM aligns segmented organ volumes with report
sentences \cite{song2025owt,yamamoto2026totalfm}. Pan-FM jointly pretrains
across seven organ and tissue MRI inputs under missing-organ settings
\cite{wu2026panfm}. Their anatomical structure arises from cross-region
training, token-group disentanglement, separated inputs, or multi-organ fusion.
OrganLens addresses a different setting: one shared encoder maps the same CT
examination to representations indexed by a selected organ identity.

\subsection{Localized and Query-Based Vision--Language Learning}
Medical vision--language models localize features through semantic queries.
GLoRIA aligns report words with image regions \cite{huang2021gloria}, MedKLIP
associates medical entities and knowledge descriptions with spatial regions
\cite{wu2023medklip}, and KAD conditions radiograph models on disease queries
\cite{zhang2023knowledge}.

CT-GLIP constructs organ-level image--text pairs, fVLM aligns anatomical
regions with report descriptions, MG-3D combines global alignment with local
reconstruction, and OKA-CT derives organ-hierarchical report knowledge for
organ-specific supervision
\cite{lin2024ctglip,shui2025large,ni2024mg3d,you2026oka}. EXACT jointly learns
organ segmentation and report-supervised anomaly localization, producing
disease-specific voxel maps constrained to the corresponding anatomy
\cite{bai2026exact}. Their objectives center on text alignment, zero-shot
diagnosis, retrieval, localization, or anomaly mapping.

Taken together, prior methods obtain anatomical specificity from masks,
isolated inputs, fixed-domain encoders, token decomposition, multi-organ
fusion, or report queries. OrganLens uses a selected organ to condition a
shared image-only encoder while the full slice remains visible. A predicted spatial mask provides weights
for feature pooling without requiring external masks at inference.

\section{Method}
\subsection{Overview and Problem Formulation}
Let $V=\{x_s\}_{s=1}^{S}$ denote a CT volume represented at inference by $S$
uniformly sampled slices that retain surrounding anatomical context, and let
$o\in\mathcal{O}$ denote an organ identity. We seek a shared foundation model
that maps the same CT volume to an organ-specific representation
$\mathbf{z}(V,o)$. During pretraining, each CT slice $x$ is paired with organ
identity $o$ and its binary mask $y^o$.

We initialize the shared encoder from a DINOv2 vision transformer
\cite{oquab2023dinov2} pretrained on CT images. As shown in
Figure~\ref{fig:organlens-overview}a, OrganLens learns organ-specific
representations through three key components. Organ-identity conditioning adds
a learned organ embedding to the CLS token, adapting encoder features to the
selected organ. Organ-guided crop sampling constructs global and local views
that retain the selected anatomy for organ-conditioned student--teacher
distillation. Anatomy-mask supervision trains an auxiliary decoder to predict
a spatial mask for the selected organ, which weights patch features within
each slice and slice representations across the CT volume. The ground-truth
mask $y^o$ is used only for crop sampling and spatial supervision and is not
passed to the encoder. At inference, only the sampled CT slices and organ
identity are required. Figure~\ref{fig:organlens-overview}b summarizes the
downstream applications.

\subsection{Organ-Identity Conditioning}
The ``organ bank'' and ``tokenizer'' in
Figure~\ref{fig:organlens-overview}a denote a learned embedding lookup that
maps a discrete organ identity to an additive CLS offset. Let $\mathbf{e}_o$
be the learned embedding for organ $o$,
$\alpha$ a trainable scalar, and $\mathbf{c}$ the learned CLS token. For input
patch embeddings $\{\mathbf{u}_j\}_{j=1}^{N}$, conditioning and encoding are
written together as
\begin{equation}
  [\mathbf{h}_{\mathrm{cls}}^o,(\mathbf{h}_j^o)_{j=1}^{N}]
  = f_\theta([\mathbf{c}+\alpha\mathbf{e}_o,
  (\mathbf{u}_j)_{j=1}^{N}]).
  \label{eq:conditioned-encoder}
\end{equation}
Initializing $\alpha$ to zero preserves the CT-pretrained model at the first
forward pass; subsequent optimization learns the strength of the organ offset.
The ``organ token'' in
Figure~\ref{fig:organlens-overview} denotes an additive offset rather than an
extra sequence token, leaving the backbone architecture unchanged. Every
augmented view and both networks receive the same organ identity, so student and
teacher representations are compared under identical conditioning.

\subsection{Organ-Specific Distillation}
Stages 2 and 4 of Figure~\ref{fig:organlens-overview}a adapt DINOv2
student--teacher training \cite{oquab2023dinov2} to the organ identity. Each
crop is constructed jointly from a CT slice and the mask corresponding to its
organ identity. Geometric transformations are applied to both, whereas
intensity augmentation affects only the image. Candidate global and local
crops must retain foreground from the anatomy specified by the organ identity.
This organ-guided cropping ensures that both student and teacher views contain the anatomy specified by the identity.

The student processes global and local crops; the teacher receives only global
crops. The teacher parameters are updated as an
exponential moving average of the corresponding student parameters,
including the backbone, organ conditioner, projection heads, and anatomy
decoder. Gradients propagate only through the student.
Following DINOv2, we combine image-level DINO distillation
\cite{caron2021emerging}, patch-level iBOT prediction \cite{zhou2022ibot}, and
KoLeo regularization
\cite{sablayrolles2019spreading,oquab2023dinov2}. Let
$\mathbf{q}_t^g(o)$ and $\mathbf{p}_s^v(o)$ denote teacher and student
prototype distributions from global view $g$ and student view $v$ under the
same organ identity.
For a masked global view $g$, let $\mathcal{M}_g$ denote its masked patch
positions, and let $k$ index the $K$ prototypes. The image- and patch-level
losses are
\begin{equation}
\begin{aligned}
  \mathcal{L}_{\mathrm{DINO}}
  &=
  -\sum_{k=1}^{K}
  q_{t,k}^{g}(o)\log p_{s,k}^{v}(o), \\
  \mathcal{L}_{\mathrm{iBOT}}
  &=
  -\frac{1}{|\mathcal{M}_g|}
  \sum_{j\in\mathcal{M}_g}\sum_{k=1}^{K}
  q_{t,j,k}^{g}(o)\log p_{s,j,k}^{g}(o).
\end{aligned}
\label{eq:self-distillation-losses}
\end{equation}
The DINO loss is averaged over each teacher global view and all student views
other than the matching global view. The iBOT loss is averaged over masked
global views.
KoLeo acts on normalized student CLS features and discourages
representation collapse. Temperature scaling, teacher centering, and loss
aggregation otherwise follow DINOv2.

\subsection{Anatomy-Mask Supervision}
Self-distillation alone does not explicitly encourage patch features to localize the queried organ. The anatomy-mask supervision stage in
Figure~\ref{fig:organlens-overview}a addresses this by predicting
organ-specific spatial mask logits from student patch features for each
global crop. The patch sequence is reshaped to its
two-dimensional grid and processed with progressive convolution and bilinear
upsampling. For sample $i$
and global view $g$, let $y_{ig}^o$ be the aligned binary anatomical target,
$\ell_{ig}^o$ the predicted logits, and
$\pi_{ig}^o=\sigma(\ell_{ig}^o)$ the predicted probability map. We combine pixel-wise binary cross-entropy with a differentiable Dice-based
loss \cite{milletari2016vnet} to address foreground imbalance. For notational brevity, let
$\pi=\pi_{ig}^o$ and $y=y_{ig}^o$ within one sample--view pair, and let $P$
denote the number of pixels. The component losses and their aggregation are
\begin{equation}
\begin{array}{@{}l@{\;=\;}l@{}}
  \mathcal{L}_{\mathrm{BCE}}(\pi,y) &
  \displaystyle -\frac{1}{P}\sum_{r=1}^{P}\left[
  y_r\log\pi_r+(1-y_r)\log(1-\pi_r)\right], \\
  \mathcal{L}_{\mathrm{Dice}}(\pi,y) &
  \displaystyle 1-\frac{2\sum_{r=1}^{P}\pi_r y_r+\epsilon_{\mathrm D}}
  {\sum_{r=1}^{P}\pi_r+\sum_{r=1}^{P}y_r+\epsilon_{\mathrm D}}, \\
  \mathcal{L}_{\mathrm{anatomy}} &
  \displaystyle \underset{i,g}{\operatorname{mean}}\left[
  \mathcal{L}_{\mathrm{BCE}}(\pi_{ig}^o,y_{ig}^o)
  +\mathcal{L}_{\mathrm{Dice}}(\pi_{ig}^o,y_{ig}^o)\right].
\end{array}
  \label{eq:anatomy-loss}
\end{equation}
For numerical stability, BCE is evaluated directly from logits. Dice is
computed per sample and global view and then averaged, with
$\epsilon_{\mathrm D}=10^{-6}$. The DINO, iBOT, and KoLeo terms follow DINOv2,
giving the complete pretraining objective
\begin{equation}
  \mathcal{L} =
  \mathcal{L}_{\mathrm{DINO}} + \mathcal{L}_{\mathrm{iBOT}}
  + \lambda_{\mathrm{KoLeo}}\mathcal{L}_{\mathrm{KoLeo}}
  + \mathcal{L}_{\mathrm{anatomy}}.
  \label{eq:total-loss}
\end{equation}
We set $\lambda_{\mathrm{KoLeo}}=0.1$.
Backpropagation through the student's anatomy decoder shapes the patch features
during pretraining. At inference, the teacher's anatomy decoder supplies the
soft weights used for anatomy-weighted pooling.

\begin{figure*}[t]
  \centering
  \includegraphics[width=\textwidth]{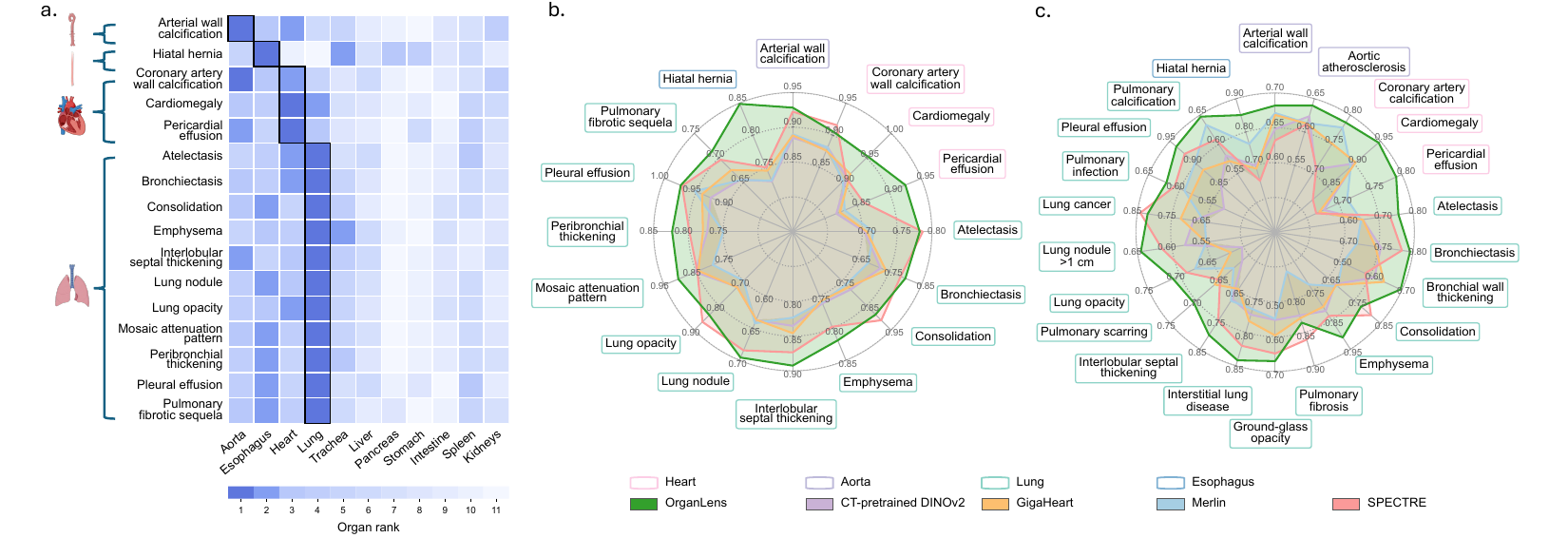}
  \caption{Organ-specific abnormality detection on CT-RATE and RAD-ChestCT.
(a) AUROC ranks of the 11 organ-specific representations for each CT-RATE
abnormality, where rank 1 indicates the best-performing organ representation.
Black outlines mark the predefined anatomically matched organ for each
abnormality. (b) AUROC by abnormality on CT-RATE and (c) RAD-ChestCT, comparing
the anatomically matched OrganLens representation with baseline models using
volume-level representations. Label-outline colors indicate the organ matched
to each abnormality. CT-RATE results are on the held-out validation split.}
  \Description{Panel a shows a heatmap of the ranks of 11 organ-specific
  representations across 18 CT-RATE abnormality labels, with outlined cells
  marking the anatomically matched representations. Panels b and c show radar
  plots comparing anatomically matched OrganLens representations with
  CT-pretrained DINOv2, GigaHeart, and Merlin on CT-RATE and RAD-ChestCT.}
  \label{fig:abnormality-results}
\end{figure*}

\begin{figure*}[t]
  \centering
  \includegraphics[width=\textwidth]{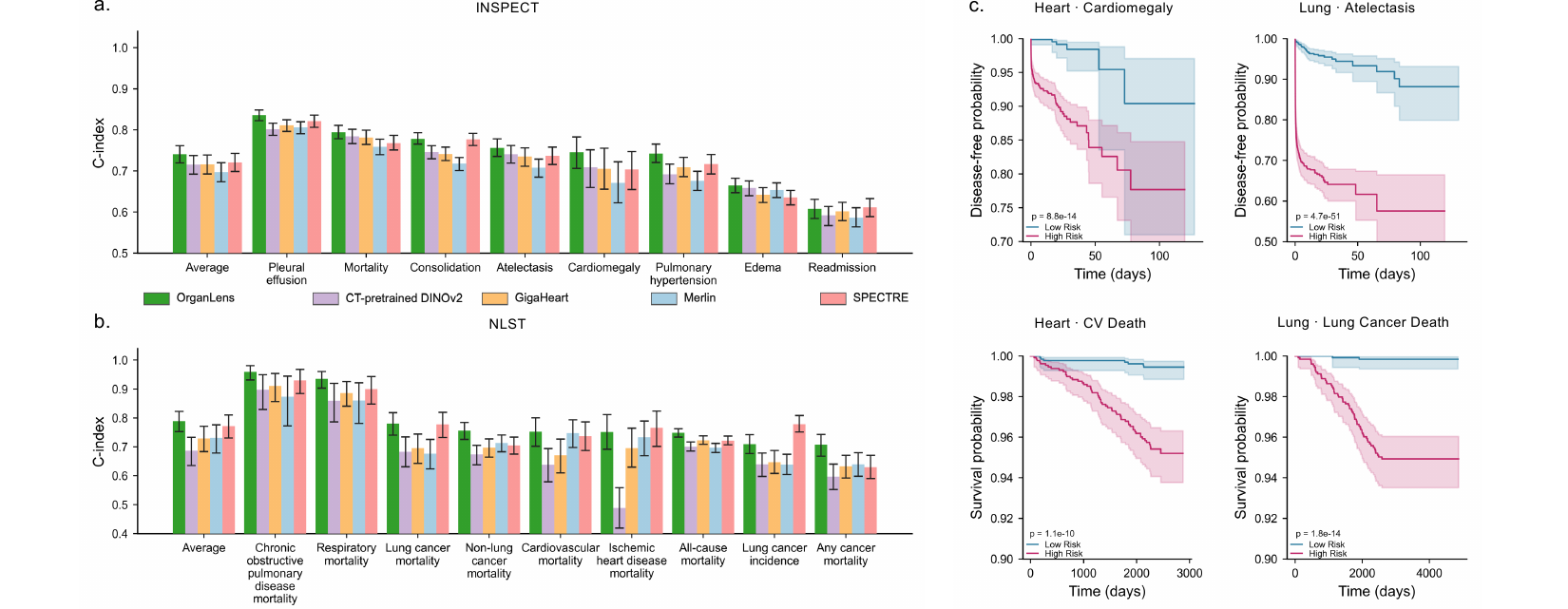}
    \caption{Organ-specific prognostic evaluation on INSPECT and NLST.
    (a,b) Test Harrell C-indices for OrganLens, CT-pretrained DINOv2,
    GigaHeart, Merlin, and SPECTRE. OrganLens results use the anatomically matched
    representation for each endpoint. Error bars are 95\%
    confidence intervals from 1,000 patient-level bootstrap resamples.
    (c) Kaplan--Meier curves compare the lowest and highest quartiles of
    held-out Cox log-hazard scores, excluding the middle 50\%.}
  \label{fig:main-results}
\end{figure*}

\begin{figure*}[t]
  \centering
  \includegraphics[width=\textwidth]{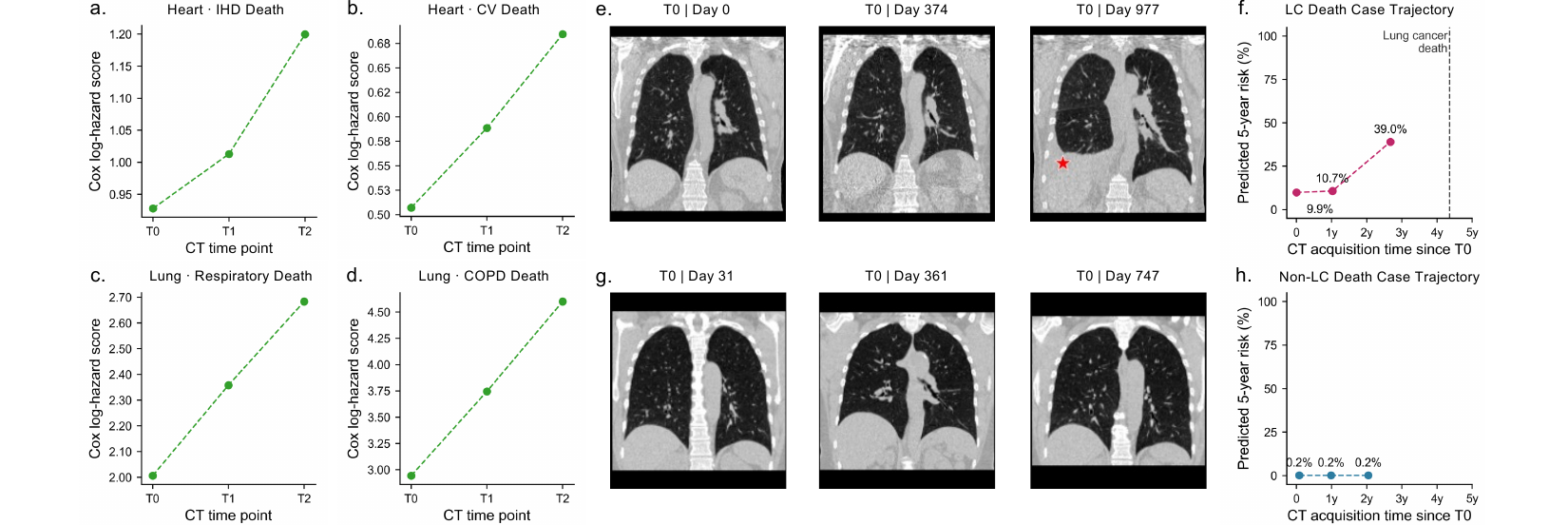}
  \caption{Longitudinal NLST assessment using OrganLens. (a--d) Mean anatomically matched Cox
  log-hazard scores at T0--T2 among participants who experienced four
  mortality endpoints. (e--h) Serial coronal CT images and 5-year
  lung-cancer-mortality estimates for a case and comparison participant. The
  red marker denotes a radiologist-identified T2 abnormality.}
  \Description{Panels a through d show increasing mean Cox log-hazard scores
  across three NLST screening time points for four cardiopulmonary mortality
  endpoints. Panels e and f show serial CT images and increasing model-derived
  5-year lung-cancer-mortality estimates for a case, with a radiologist-marked
  abnormality at T2. Panels g and h show a comparison participant with stable
  low estimates.}
  \label{fig:longitudinal-results}
\end{figure*}

\begin{figure}[t]
  \centering
  \includegraphics[width=\columnwidth]{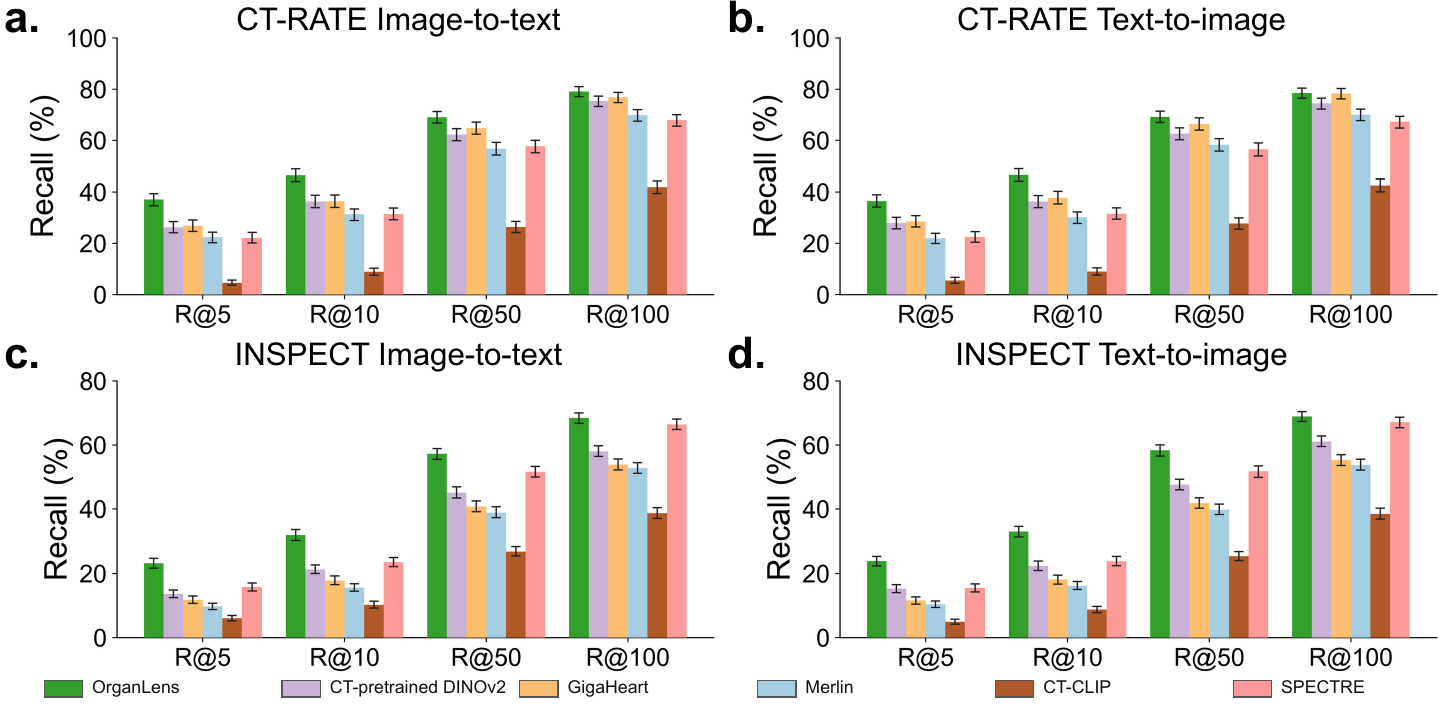}
  \caption{Image--text retrieval on CT-RATE (a,b) and INSPECT (c,d).
  Bars show Recall@$K$ with 95\% confidence intervals from 1,000 bootstrap
resamples with candidate sets held fixed.}
  \Description{Four grouped bar plots compare OrganLens, CT-pretrained DINOv2,
  GigaHeart, Merlin, CT-CLIP, and SPECTRE for image-to-text and text-to-image
  retrieval on CT-RATE and INSPECT.}
  \label{fig:retrieval-results}
\end{figure}

\subsection{Organ-Specific Representation}
\label{sec:organ-specific-representation}
At inference, each slice $x_s$ is encoded under organ identity $o$.
The encoder returns an organ-specific CLS feature
$\mathbf{h}_{s,\mathrm{cls}}^o$ and patch features
$\{\mathbf{h}_{sj}^o\}_{j=1}^{N}$. The anatomy decoder produces spatial mask
logits from the patch features. The CLS feature carries the organ-identity
conditioning and supports image-level self-distillation and KoLeo
regularization during pretraining. The final organ-specific representation is
constructed from spatially pooled patch features. We apply a sigmoid to the
logits and average-pool the probabilities to the patch grid, obtaining soft
weights $m_{sj}^o\in[0,1]$. For every sampled slice, the slice representation
and predicted organ area are
\begin{equation}
  \mathbf{z}_s^o =
  \frac{\sum_{j=1}^{N}m_{sj}^o\mathbf{h}_{sj}^o}
       {\sum_{j=1}^{N}m_{sj}^o},
  \qquad
  a_s^o = \frac{1}{N}\sum_{j=1}^{N}m_{sj}^o.
  \label{eq:slice-pooling}
\end{equation}
We retain all $S$ sampled slices and aggregate their representations in
proportion to predicted organ area:
\begin{equation}
  \mathbf{z}(V,o) =
  \frac{\sum_{s=1}^{S}a_s^o\mathbf{z}_s^o}
       {\sum_{s=1}^{S}a_s^o}.
  \label{eq:volume-pooling}
\end{equation}
Applying this procedure to each organ identity produces one
organ-specific representation from the same CT volume using shared model
parameters.

Downstream tasks may use any organ-specific representation or the global
representation defined below. For tasks requiring whole-volume information, let
$\mathcal{O}=\{o_1,\ldots,o_{11}\}$ denote the 11 organ identities used by
OrganLens: heart, lung, aorta, liver, spleen, kidneys, pancreas, stomach,
intestine, esophagus, and trachea.
We define the global representation by concatenating all 11
organ-specific representations:
\begin{equation}
  \mathbf{z}_{\mathrm{global}}(V) =
  \mathbf{z}(V,o_1)\,\Vert\,\cdots\,\Vert\,\mathbf{z}(V,o_{11}),
  \label{eq:global-representation}
\end{equation}
where $\Vert$ denotes feature concatenation. This construction adds no
pretraining module. The downstream head operates on the concatenated features.
Each organ identity requires one conditioned encoder pass, so an organ-specific
task needs one pass and the global representation needs 11, which can be
batched.
The retrieval experiments use all 11 organ identities so that CT-RATE and INSPECT
use the same image representation.

\section{Data and Experimental Setup}
\subsection{Study Design and Cohorts}
For OrganLens, CT-RATE is the sole adaptation cohort and also supports internal
evaluation. Its validation split was not used for pretraining, hyperparameter
selection, or checkpoint selection and served as the held-out internal test set. RAD-ChestCT, INSPECT, and NLST are used only for downstream
transfer. Across the four cohorts, abnormality labels, clinical outcomes, serial scans,
and reports support the four evaluation task families.

\noindent\textbf{CT-RATE.}
CT-RATE includes data from 21,304 patients and a total of 50,188
non-contrast 3D chest CT volumes, including alternative reconstructions. The images
support pretraining, while radiology reports, multi-abnormality labels, and
acquisition metadata support abnormality detection and image--text
retrieval~\cite{hamamci2024developing}.

\noindent\textbf{RAD-ChestCT.}
The public RAD-ChestCT release used here contains 3,630 scans with 23
abnormality labels automatically extracted from reports~\cite{draelos2021machine,draelos_2020_6406114}. It contains
2,286 training, 984 validation, and 360 held-out test volumes and evaluates
supervised cross-cohort transfer rather than zero-shot generalization.

\noindent\textbf{INSPECT.}
INSPECT contains 23,248 CT pulmonary angiography studies from 19,402 patients,
with 3D images, report impressions, longitudinal electronic health records,
and diagnostic and prognostic labels~\cite{huang2023inspect}. It tests transfer
beyond non-contrast CT. Paired reports and clinical outcomes support retrieval
and prognostic prediction, respectively.

\noindent\textbf{NLST.}
NLST enrolled 53,454 participants at high risk for lung cancer and offered a
baseline plus two annual low-dose CT or chest-radiography screenings
\cite{national2011national}. We use serial CT examinations and linked cancer
and mortality records for time-to-event prediction and longitudinal analysis.

\subsection{Preprocessing and Organ Definitions}
Following the CT-CLIP-style pipeline~\cite{hamamci2024developing}, images are
resampled to $1\,$mm isotropic spacing, clamped to $[-1000,1000]$ HU, and
scaled to $[-1,1]$. Pretraining uses aligned axial, coronal, and sagittal
image--mask pairs. Following GigaHeart~\cite{xu2025cardiac}, we represent each
downstream volume with $S=64$ uniformly sampled axial slices, each center
cropped to $224\times224$. Supplementary
Section~\ref{sec:detailed-preprocessing} provides the interpolation,
crop--padding, channel-conversion, and normalization details.
Each pretraining unit contains a CT slice, an organ identity, and its binary
mask. We apply TotalSegmentator~\cite{wasserthal2023totalsegmentator} to
CT-RATE and use the resulting pseudo-label masks to define 11 targets: heart,
lung, aorta, liver, spleen, kidneys, pancreas, stomach, intestine, esophagus,
and trachea. Bilateral kidneys, lung lobes, and intestinal substructures are
merged before target selection.

\subsection{Pretraining and Implementation Details}
We initialize the student and teacher ViT-L/16 backbones
\cite{dosovitskiy2021image} from the CT-pretrained checkpoint released with
GigaHeart \cite{xu2025cardiac}, which was trained on an external chest CT
collection using the DINOv2 framework \cite{oquab2023dinov2}.
Each sample produces two $224\times224$ global crops and eight $96\times96$
local crops using organ-guided sampling. Pretraining uses AdamW for three
complete passes and 525,000 optimization iterations with a global batch size
of 1,024 across eight GPUs and a base learning rate of $2\times10^{-4}$.
The recipe follows DINOv2, the architecture and initialization follow
GigaHeart, and the OrganLens-specific settings were fixed a priori.
Supplementary Section~\ref{sec:pretraining-details} provides the crop-acceptance and
fallback rules, iBOT masking settings, optimization schedules, decoder
configuration, and numerical details.
Generative AI usage is described in
Section~\ref{sec:generative-ai-usage}.

\section{Downstream Evaluation and Results}
Supplementary Section~\ref{sec:downstream-training-details} provides downstream
architectures, optimization settings, and model-selection procedures. Across downstream tasks, we compare OrganLens with CT-pretrained DINOv2, 
GigaHeart~\cite{xu2025cardiac}, Merlin~\cite{blankemeier2026merlin}, and
SPECTRE~\cite{claessens2025spectre}. Image--text retrieval additionally
includes CT-CLIP~\cite{hamamci2024developing}. Available baselines follow the
same task-specific data splits and downstream protocols.
SPECTRE's pretraining data included the CT-RATE training split, INSPECT, and NLST. Its
evaluations on these cohorts are therefore pretraining-exposed
rather than fully external transfer \cite{claessens2025spectre}.
For analyses requiring a global OrganLens representation, the main results
use concatenation. Corresponding mean aggregation results are reported in the
supplementary tables.

\subsection{Abnormality Detection}
We evaluated abnormality detection across 16 CT-RATE abnormalities and 22
RAD-ChestCT abnormalities. To assess the quality of the learned representations, we extracted representations using frozen encoders and trained a separate lightweight
one-hidden-layer MLP probe for each abnormality. The CT-RATE probes were trained for a fixed five epochs without checkpoint selection and evaluated on the held-out validation split,
whereas RAD-ChestCT performance was reported on the held-out test split. 
For nearly all abnormalities, the anatomically matched representation achieved
the highest performance among the OrganLens representations (Figure~\ref{fig:abnormality-results}a). Supplementary
Table~\ref{tab:abnormality-organ-auroc} reports the complete matrices. 

Across 16 CT-RATE abnormalities, anatomically matched OrganLens
representations achieved a macro-AUROC of 0.856
(Figure~\ref{fig:abnormality-results}b), compared with 0.788, 0.792, 0.780,
and 0.836 for CT-pretrained DINOv2, GigaHeart, Merlin, and SPECTRE,
respectively. OrganLens achieved the highest AUROC for 12 of the 16
abnormalities.
For cardiomegaly and pericardial effusion, the heart representation reached
AUROCs of 0.953 and 0.925, outperforming the strongest baselines, GigaHeart
at 0.917 and SPECTRE at 0.860, respectively. The esophagus representation
reached 0.849 for hiatal hernia, compared with 0.750 for SPECTRE, while the
lung representation reached 0.906 for mosaic attenuation pattern, compared
with 0.853 for CT-pretrained DINOv2.

On the external RAD-ChestCT dataset, anatomically matched OrganLens
representations achieved a macro-AUROC of 0.767
(Figure~\ref{fig:abnormality-results}c), compared with 0.684, 0.694, 0.688,
and 0.721 for CT-pretrained DINOv2, GigaHeart, Merlin, and SPECTRE,
respectively. OrganLens achieved the highest AUROC for 19 of 22 abnormalities.
Across 1,000 paired bootstrap resamples, OrganLens reached macro-AUROCs of
0.856 [0.848, 0.863] on CT-RATE and 0.767 [0.751, 0.782] on RAD-ChestCT, with
95\% confidence intervals in brackets, and exceeded SPECTRE by 0.020 [0.014,
0.026] and 0.046 [0.030, 0.061]. It exceeded every baseline on both cohorts
across five probe seeds.
Its advantages were preserved for cardiomegaly, reaching 0.946 compared with
0.903 for GigaHeart, and for pericardial effusion, reaching 0.791 compared
with 0.714 for Merlin. OrganLens performed strongly on abnormalities not
included in CT-RATE evaluation, including interstitial lung disease, for which
the lung representation reached 0.842 compared with 0.820 for SPECTRE. These
results support the cross-cohort generalizability of organ-specific
representations.

\subsection{Prognostic Prediction}
We assessed prognostic prediction across eight INSPECT endpoints and nine
NLST endpoints. To assess the quality of the learned representations, we
extracted representations using frozen encoders and trained a
lightweight MLP Cox model for each endpoint. INSPECT used its
predefined splits, whereas NLST used a patient-level 70\%/10\%/20\% split.  Cardiovascular
endpoints used the heart representation, pulmonary and lung-cancer endpoints
used lung representation, and broader outcomes used global representations.
Supplementary Table~\ref{tab:cox-cindex} reports all results.

Across eight INSPECT endpoints, the anatomically matched representation
achieved the highest C-index among all OrganLens representations for seven
endpoints and a mean C-index of 0.741
(Figure~\ref{fig:main-results}a), compared with 0.715 for CT-pretrained DINOv2,
0.716 for GigaHeart, 0.697 for Merlin, and 0.721 for SPECTRE. The
largest gains over the strongest endpoint-specific baseline were 0.037 for
cardiomegaly, 0.026 for pulmonary hypertension, and 0.015 for atelectasis.

Across nine NLST endpoints, matched OrganLens representations achieved the
highest C-index for seven and a mean C-index of 0.789
(Figure~\ref{fig:main-results}b). CT-pretrained DINOv2, GigaHeart, Merlin, and
SPECTRE averaged 0.686, 0.729, 0.731, and 0.772. SPECTRE was pretrained on the
full NLST cohort~\cite{claessens2025spectre}. For lung-cancer, respiratory, and
COPD mortality, the lung representation reached C-indices of 0.780, 0.935, and
0.960, exceeding the strongest baseline, SPECTRE, by 0.003, 0.035, and 0.030.
In a paired patient-level bootstrap on patients scored by all models, the mean
C-index difference from SPECTRE was 0.019 [0.011, 0.028] on INSPECT and 0.016
[$-0.003$, 0.036] on NLST, and all intervals against the other baselines
excluded zero.

Kaplan--Meier curves compared the lowest and highest risk quartiles
(Figure~\ref{fig:main-results}c), with log-rank $p$ values reported in each
panel. Across the displayed endpoints, the high-risk quartiles showed
consistently lower event-free survival, supporting the prognostic
stratification captured by anatomically matched representations. These curves
assess stratification, not calibration.

\subsection{Longitudinal Analysis}
The longitudinal analysis used serial CT examinations and linked mortality
follow-up from NLST. Participants with fewer than three complete screening
time points formed the training and validation cohorts, which used all of their
available scans. Participants with CT examinations available at all three time points (T0, T1, and T2) comprised the longitudinal test cohort. At each test visit, we used the scan from that visit
to estimate mortality risk over the following five years. This is a landmark
design, with the clock restarted at each visit. Each visit was
evaluated independently, and the Cox risk score was converted to an absolute
risk using the baseline hazard estimated from the training cohort.

Across the common longitudinal test cohort, OrganLens achieved the highest
C-index point estimate in 11 of 12 endpoint-visit comparisons across the T0,
T1, and T2 screening visits, with COPD mortality at T0 comparable to SPECTRE.
Comparisons for each endpoint across the T0, T1, and T2 visits are provided in
Supplementary Figure~\ref{fig:longitudinal-baselines}.

Among participants who later experienced each endpoint, mean anatomically
matched Cox scores increased monotonically from T0 to T2
(Figure~\ref{fig:longitudinal-results}a--d). Heart representation scores at T0, T1, and T2
were 0.928, 1.013, and 1.200 for ischemic-heart-disease mortality ($n=150$),
and 0.507, 0.589, and 0.685 for cardiovascular mortality ($n=278$). Lung representation
scores were 2.006, 2.358, and 2.684 for respiratory mortality ($n=120$), and
2.944, 3.745, and 4.600 for COPD mortality ($n=79$). The four
endpoint-specific trajectories were not on a common score scale.
Scores of participants alive through four years after T2 also rose. Among
deaths within four years after T2, the increase from T0 to T2 exceeded that of
these survivors by 0.29 [0.15, 0.43] for respiratory and 1.03 [0.67, 1.38] for
COPD mortality, with no significant difference for ischemic-heart-disease and
cardiovascular mortality (Supplementary Figure~\ref{fig:longitudinal-controls}).

For a participant who died of lung cancer on day 1,590, uncalibrated model-derived 5-year
mortality estimates were 9.9\%, 10.7\%, and 39.0\% on days 0, 374, and 977
(Figure~\ref{fig:longitudinal-results}e,f). A radiologist-marked T2 abnormality
coincided with the sharp final-visit risk increase. For comparison, a participant who did not die from lung cancer had estimated risks of 0.16\%, 0.17\%, and 0.17\% on
days 31, 361, and 747
(Figure~\ref{fig:longitudinal-results}g,h). Each estimate used the scan from the corresponding visit and was not
accumulated from T0.

\subsection{Image--Text Retrieval}
Retrieval used the scan-level global representation defined in Section~\ref{sec:organ-specific-representation}. For every model evaluated, the image encoder, ClinicalBERT text encoder
\cite{clinicalbert}, and projection heads were jointly
fine-tuned for five epochs using the symmetric CLIP objective
\cite{radford2021learning}. CT-RATE text inputs combined the report findings
and impression, whereas INSPECT text inputs used the report impression.

We evaluated Recall@$K$ in both directions with one paired item per identity.
The official evaluation splits contained 1,564 CT-RATE and 3,212 INSPECT
image--text pairs. All methods used the same fixed candidate set within each
cohort. OrganLens outperformed all evaluated baselines at every $K$ in both
text-to-image and image-to-text retrieval on both cohorts
(Figure~\ref{fig:retrieval-results}). On CT-RATE, text-to-image and
image-to-text Recall@5 values were 36.45\% and 37.08\%, while Recall@10 values
were 46.55\% and 46.42\%. On INSPECT, the corresponding values were 23.88\%
and 23.19\% at Recall@5 and 33.09\% and 32.04\% at Recall@10.
Mean aggregation, which has the same width as the CT-pretrained DINOv2
representation, reached text-to-image Recall@5 values of 37.47\% on CT-RATE
and 22.63\% on INSPECT, compared with 27.88\% and 15.29\% for CT-pretrained
DINOv2 and 15.47\% for SPECTRE on INSPECT.

\subsection{Ablation Study}

\begin{figure}[t]
  \centering
  \includegraphics[width=\columnwidth]{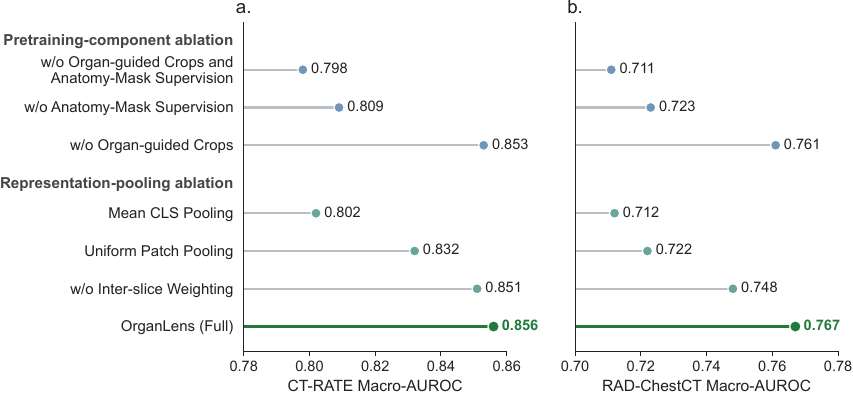}
  \caption{Pretraining-component and representation-pooling ablations on
(a) CT-RATE and (b) RAD-ChestCT. Points show macro-AUROC for each variant.
OrganLens (Full), highlighted in green, is the common reference for both
ablations.}
  \Description{Two horizontal lollipop plots show macro-AUROC for three
  pretraining-component ablations and four representation-pooling variants on
  CT-RATE and RAD-ChestCT. OrganLens Full is highlighted in green.}
  \label{fig:ablation-results}
\end{figure}

We ablated organ-guided cropping and anatomy-mask supervision in a two-by-two
design. Variants without anatomy-mask supervision used mean CLS pooling,
whereas variants with it used anatomy-weighted pooling. Removing both
components yielded macro-AUROCs of 0.798 on CT-RATE and 0.711 on RAD-ChestCT.
Removing anatomy-mask supervision yielded 0.809 and 0.723. Removing
organ-guided crops yielded 0.853 and 0.761. OrganLens (Full) reached 0.856 and
0.767 (Figure~\ref{fig:ablation-results}a,b).
Removing only the organ offset in Eq.~(\ref{eq:conditioned-encoder}) while
retaining organ-guided crops and anatomy-mask supervision yielded 0.818 on
CT-RATE and 0.685 on RAD-ChestCT with anatomy-weighted pooling, and 0.817 and
0.718 with mean CLS pooling. An adaptation-only variant without organ-identity
conditioning, organ-guided crops, or anatomy-mask supervision yielded 0.797
and 0.718, compared with 0.792 and 0.694 for GigaHeart. With mean CLS pooling
for both, the variants without and with anatomy-mask supervision yielded 0.809
and 0.802 on CT-RATE and 0.723 and 0.712 on RAD-ChestCT. Pooling CT-pretrained
DINOv2 and GigaHeart features with TotalSegmentator masks of the matched organ
yielded 0.809 and 0.822 on CT-RATE and 0.718 for both on RAD-ChestCT. The
matched organ representation ranked first among the 11 organ representations
for 15 of 16 CT-RATE and 18 of 22 RAD-ChestCT abnormalities. Averaged over
abnormalities, the ten mismatched organs reached 0.815 and 0.695, and the best
mismatched organ per abnormality reached 0.835 and 0.743.

We next ablated the two anatomy-weighted pooling stages while retaining the
organ-specific encoder. Mean CLS pooling averaged organ-specific slice
CLS features, while uniform patch pooling averaged unweighted patch features.
The variant without inter-slice weighting retained predicted within-slice
spatial mask but averaged slices uniformly, whereas OrganLens (Full)
weighted slices by predicted organ area. Complete label-wise results for
matched, mean, and concatenated representations are reported in Supplementary
Tables~\ref{tab:pretraining-ablation} and~\ref{tab:pooling-ablation}.

Mean CLS pooling achieved macro-AUROCs of 0.802 on CT-RATE and 0.712 on
RAD-ChestCT. Uniform patch pooling increased them to 0.832 and 0.722. Removing
inter-slice weighting yielded 0.851 and 0.748. OrganLens (Full) reached 0.856
and 0.767 (Figure~\ref{fig:ablation-results}a,b), improving 12 of 16 CT-RATE
abnormalities, with two ties, and 19 of 22 RAD-ChestCT abnormalities relative
to uniform slice weighting. Overall, applying predicted spatial mask within
slices and weighting slices by predicted organ area both contributed, with
larger gains from the former on CT-RATE and from both stages on RAD-ChestCT.

\section{Conclusion}
OrganLens addresses the mismatch between volume-level CT representations and
organ-focused biomedical questions through a shared encoder conditioned on an
organ identity. Anatomy-mask supervision shapes patch features for pooling with a 
predicted spatial mask, removing the need for external segmentation masks
at inference. Across diverse cohorts and task families, anatomically matched
representations generally improved abnormality detection, prognostic
prediction, and longitudinal risk assessment over volume-level baselines. 
The global representation preserved broader transfer and achieved the
strongest image--text retrieval performance on both evaluated cohorts.
Consistent cross-cohort performance further supports the generalizability of
OrganLens across acquisition settings and downstream tasks. OrganLens therefore
offers a scalable framework for studying organ-specific phenotypes,
longitudinal change, and clinical risk across CT cohorts.

\section{Limitations and Ethical Considerations}
Although OrganLens was retrospectively evaluated across four chest CT cohorts,
OrganLens adaptation was limited to CT-RATE and may therefore reflect
cohort-specific biases. Its prospective performance and
transfer beyond chest CT remain untested, and adaptation across additional
domains is future work. TotalSegmentator pseudo-label errors may
propagate into the predicted spatial mask
(Supplementary Section~\ref{sec:mask-quality}). Two-dimensional encoding of 64
sampled slices does not model volumetric continuity and may miss small or
sparsely sampled findings. Explicit inter-slice modeling is a direction for
such findings. Each ablation variant was pretrained once, which limits
conclusions about small differences. 

Reported discrimination and retrieval metrics do
not establish clinical benefit. The prognostic models were not assessed for
calibration, and longitudinal predictions were repeated cross-sectional
estimates rather than dynamic updates. Data were deidentified and accessed
under the dataset agreements, and no reidentification was attempted.
OrganLens should not guide care without prospective validation, uncertainty
assessment, and clinician oversight.

\section{Generative AI Usage}
\label{sec:generative-ai-usage}
Generative AI tools assisted with experimental code development and debugging,
data analysis, figure and table preparation, literature organization, and
manuscript editing. All AI-assisted outputs were reviewed and verified by the
authors, who retain full responsibility for the work.

\clearpage
\bibliographystyle{ACM-Reference-Format}
\bibliography{references}

\appendix

\setcounter{table}{0}
\renewcommand{\thetable}{S\arabic{table}}
\renewcommand{\tablename}{Supplementary Table}
\setcounter{figure}{0}
\renewcommand{\thefigure}{S\arabic{figure}}
\renewcommand{\figurename}{Supplementary Figure}

\section{Additional Reproducibility Details}
\subsection{Shared Downstream Protocol}
Unless stated otherwise, downstream evaluations used frozen teacher features,
with all examinations from each patient assigned to one split. The CT-RATE
validation split was not used for hyperparameter or checkpoint selection, and
available baselines
followed the same splits and downstream protocol. Detection confidence
intervals resample patients on CT-RATE and volumes on RAD-ChestCT, which lacks
patient identifiers. Every organ-specific
representation was computed from 64 uniformly sampled axial slices.

\subsection{Detailed Image and Mask Preprocessing}
\label{sec:detailed-preprocessing}
Images and segmentation masks are transformed on the same spatial grid using
data-appropriate interpolation. Images are resampled to $1\,$mm isotropic
spacing with trilinear interpolation, center cropped or padded to $512^3$
voxels, clamped to $[-1000,1000]$ HU, and scaled to $[-1,1]$. Padded voxels
take the value $-1$. Segmentation volumes use nearest-neighbor interpolation
and zero padding to preserve discrete labels. We extract aligned axial,
coronal, and sagittal image--mask pairs at $256\times256$ resolution, map each
image slice to three-channel grayscale, and apply DINO normalization after
spatial cropping. Downstream slices are center cropped to $224\times224$
before normalization.

\subsection{Pretraining Hyperparameters}
\label{sec:pretraining-details}
OrganLens was adapted on the CT-RATE training split using CT images and
TotalSegmentator pseudo-labels. Only the ViT-L/16 backbones of the student and teacher were initialized from
the CT-pretrained DINOv2 checkpoint released with
GigaHeart~\cite{xu2025cardiac}. The shared DINO/iBOT projection head,
organ-conditioning module, and anatomy decoder were initialized from scratch.

The two $224\times224$ global crops and eight $96\times96$ local crops use
scale ranges of $[0.32,1.0]$ and $[0.05,0.32]$, respectively. A candidate
global crop is accepted when it contains at least 32 organ-mask pixels, 1\%
foreground, and 75\% of the organ bounding box. A candidate local crop
requires at least 8 organ-mask pixels and 3\% foreground. If repeated sampling
fails, the crop is drawn around the organ bounding box or a foreground point.
For iBOT, masking is applied to 50\% of samples, with the patch ratio drawn
from $[0.1,0.5]$.

Each of the three complete pretraining passes contains 175,000 iterations.
The learning-rate schedule includes a 12,500-iteration warmup. Weight decay
increases from 0.04 to 0.4, and teacher momentum increases from 0.992 to 1.0. The
anatomy decoder uses progressive convolutional blocks with channel widths of
256, 128, 64, and 32, followed by bilinear upsampling and a $1\times1$
prediction layer. Training uses mixed precision and fully sharded data parallelism with a random
seed of 0. Numerical stabilizers are $10^{-8}$ for KoLeo
and $10^{-6}$ for Dice. Pretraining was performed on eight NVIDIA RTX PRO 6000 GPUs.

\subsection{Downstream Model Training}
\label{sec:downstream-training-details}
Each abnormality head is a one-hidden-layer, 512-unit MLP. The CT-RATE heads
are trained separately for each abnormality for five epochs with unweighted binary cross-entropy.
They use AdamW with a learning rate of $2\times10^{-3}$, weight decay of 0.05,
a batch size of 16, one warmup epoch, and no dropout. The RAD-ChestCT heads use
the same architecture, optimizer, learning rate, and weight decay, with a
batch size of 64 and class-weighted binary cross-entropy. They are trained for
100 epochs, with checkpoint selection performed separately for each label
using validation AUROC.

The OrganLens, CT-pretrained DINOv2, GigaHeart, and Merlin Cox models standardize
their input features and use a one-hidden-layer MLP
with 512 ReLU units, no dropout, and one scalar log-hazard output. Full-batch
optimization uses AdamW with a learning rate of $2\times10^{-3}$, weight decay
of 0.05, five warmup epochs, and cosine decay. Training runs for up to 200
epochs and stops after 20 consecutive epochs without validation Harrell
C-index improvement. These settings are fixed across representation--endpoint
pairs rather than selected through a separate hyperparameter search.

For jointly fine-tuned image--text retrieval models, the image encoder,
ClinicalBERT text encoder, and linear projection heads are optimized for five
epochs using AdamW and symmetric contrastive loss. CT-RATE uses learning rates of $10^{-5}$ for both
the encoders and projection heads, whereas INSPECT uses $10^{-5}$ for the
encoders and $10^{-4}$ for the projection heads. Both use a weight decay of 0.01.
Both runs use four GPUs and a micro-batch size of 4. CT-RATE accumulates 16
micro-batches per rank and INSPECT accumulates 32, producing local contrastive
pools of 64 and 128 pairs, respectively, before gradient averaging across
ranks. INSPECT selects the checkpoint with the highest validation text-to-image
Recall@10, whereas CT-RATE uses the fifth-epoch checkpoint for evaluation on
its validation candidate set. When multiple volume files correspond to the
same CT-RATE scan, we retain the first volume.

\section{Complete Organ-Specific Abnormality Detection Results}
\label{sec:complete-abnormality-results}
Supplementary Table~\ref{tab:abnormality-organ-auroc} reports AUROCs for every
organ-specific OrganLens representation and abnormality on the CT-RATE
validation split and RAD-ChestCT held-out test split. The table also includes
volume-level CT-pretrained
DINOv2, GigaHeart, Merlin, and SPECTRE baselines. CT-pretrained DINOv2 and
GigaHeart use mean-pooled slice-level CLS features, Merlin uses its global
image embedding, and SPECTRE uses its scan-level CLS feature. All OrganLens
representations are computed from 64 uniformly sampled axial slices and use
anatomy-weighted pooling.

\begin{table*}[t]
\centering
\caption{Complete organ-specific abnormality detection results. Panel A reports AUROCs on the held-out CT-RATE validation split, and Panel B reports held-out test AUROCs on RAD-ChestCT. Best and second-best values within each abnormality are shown in bold and underlined, respectively. Ties after rounding to three decimal places receive the same formatting. A dagger marks the anatomically matched OrganLens representation used in Figure~\ref{fig:abnormality-results}. Concat and mean aggregate all 11 organ-specific representations. CT-pretrained DINOv2 and GigaHeart use mean-pooled slice-level CLS features, Merlin uses its global image embedding, and SPECTRE uses its scan-level CLS feature. SPECTRE was trained on the CT-RATE training split.}
\label{tab:abnormality-organ-auroc}
\begingroup
\setlength{\tabcolsep}{3pt}
\renewcommand{\arraystretch}{0.96}
\resizebox{\textwidth}{!}{%
\begin{tabular}{l*{17}{c}}
\toprule
& \multicolumn{11}{c}{OrganLens organ-specific representations} & \multicolumn{2}{c}{OrganLens global aggregations} & \multicolumn{4}{c}{Volume-level baselines} \\
\cmidrule(lr){2-12}\cmidrule(lr){13-14}\cmidrule(lr){15-18}
Abnormality & Heart & Lung & Aorta & Liver & Spleen & Kidneys & Pancreas & Stomach & Intestine & Esophagus & Trachea & Concat & Mean & CT-pretrained DINOv2 & GigaHeart & Merlin & SPECTRE \\
\midrule
\multicolumn{18}{l}{\textbf{(A) CT-RATE validation}} \\
Medical material & 0.863 & 0.852 & \underline{0.885} & 0.822 & 0.849 & 0.857 & 0.811 & 0.824 & 0.818 & 0.854 & 0.847 & \textbf{0.887} & 0.874 & 0.792 & 0.804 & 0.864 & 0.873 \\
Arterial wall calcification & 0.919 & 0.915 & \textbf{0.928}\textsuperscript{$\dagger$} & 0.909 & 0.911 & 0.919 & 0.906 & 0.902 & 0.910 & 0.919 & 0.915 & 0.921 & \underline{0.922} & 0.885 & 0.888 & 0.890 & \underline{0.922} \\
Cardiomegaly & \textbf{0.953}\textsuperscript{$\dagger$} & 0.947 & 0.947 & 0.919 & 0.938 & 0.938 & 0.916 & 0.916 & 0.908 & 0.942 & 0.934 & \underline{0.950} & 0.949 & 0.910 & 0.917 & 0.902 & 0.909 \\
Pericardial effusion & \textbf{0.925}\textsuperscript{$\dagger$} & 0.880 & 0.889 & 0.844 & 0.877 & 0.857 & 0.839 & 0.864 & 0.841 & 0.871 & 0.856 & \underline{0.910} & 0.897 & 0.818 & 0.823 & 0.831 & 0.860 \\
Coronary artery wall calcification & 0.905\textsuperscript{$\dagger$} & 0.899 & \underline{0.911} & 0.898 & 0.897 & 0.902 & 0.893 & 0.890 & 0.894 & 0.904 & 0.897 & 0.905 & 0.909 & 0.881 & 0.876 & 0.881 & \textbf{0.916} \\
Hiatal hernia & 0.769 & 0.763 & 0.790 & 0.788 & 0.789 & 0.770 & 0.822 & 0.814 & 0.771 & \textbf{0.849}\textsuperscript{$\dagger$} & 0.825 & \underline{0.846} & 0.833 & 0.728 & 0.745 & 0.730 & 0.750 \\
Lymphadenopathy & 0.739 & \underline{0.753} & 0.744 & 0.727 & 0.731 & 0.727 & 0.714 & 0.720 & 0.727 & 0.749 & 0.751 & \textbf{0.756} & \underline{0.753} & 0.709 & 0.717 & 0.719 & 0.725 \\
Emphysema & 0.799 & \textbf{0.819}\textsuperscript{$\dagger$} & 0.797 & 0.791 & 0.787 & 0.788 & 0.783 & 0.775 & 0.789 & 0.801 & 0.801 & 0.804 & \underline{0.807} & 0.761 & 0.757 & 0.760 & 0.798 \\
Atelectasis & 0.758 & \underline{0.782}\textsuperscript{$\dagger$} & 0.747 & 0.743 & 0.752 & 0.728 & 0.711 & 0.727 & 0.728 & 0.749 & 0.732 & 0.765 & 0.763 & 0.703 & 0.707 & 0.716 & \textbf{0.787} \\
Lung nodule & 0.687 & \textbf{0.696}\textsuperscript{$\dagger$} & 0.685 & 0.675 & 0.676 & 0.658 & 0.642 & 0.656 & 0.670 & 0.691 & 0.679 & 0.647 & \underline{0.694} & 0.638 & 0.637 & 0.642 & 0.685 \\
Lung opacity & 0.781 & \underline{0.839}\textsuperscript{$\dagger$} & 0.777 & 0.726 & 0.753 & 0.723 & 0.681 & 0.716 & 0.742 & 0.777 & 0.765 & 0.824 & 0.816 & 0.725 & 0.726 & 0.699 & \textbf{0.867} \\
Pulmonary fibrotic sequela & 0.694 & \textbf{0.714}\textsuperscript{$\dagger$} & 0.694 & 0.674 & 0.686 & 0.677 & 0.674 & 0.668 & 0.670 & 0.695 & 0.680 & \underline{0.700} & 0.694 & 0.656 & 0.675 & 0.655 & 0.696 \\
Pleural effusion & 0.961 & \textbf{0.974}\textsuperscript{$\dagger$} & 0.961 & 0.960 & 0.962 & 0.950 & 0.939 & 0.952 & 0.934 & 0.963 & 0.958 & \underline{0.972} & 0.967 & 0.928 & 0.941 & 0.952 & \underline{0.972} \\
Mosaic attenuation pattern & 0.874 & \textbf{0.906}\textsuperscript{$\dagger$} & 0.881 & 0.861 & 0.867 & 0.854 & 0.821 & 0.839 & 0.852 & 0.882 & 0.872 & \underline{0.897} & 0.885 & 0.853 & 0.844 & 0.799 & 0.847 \\
Peribronchial thickening & 0.804 & \underline{0.823}\textsuperscript{$\dagger$} & 0.808 & 0.791 & 0.798 & 0.796 & 0.776 & 0.785 & 0.789 & 0.812 & 0.810 & \textbf{0.825} & \textbf{0.825} & 0.773 & 0.779 & 0.750 & 0.799 \\
Consolidation & 0.811 & \underline{0.887}\textsuperscript{$\dagger$} & 0.823 & 0.793 & 0.803 & 0.794 & 0.755 & 0.789 & 0.801 & 0.827 & 0.822 & 0.854 & 0.857 & 0.785 & 0.776 & 0.765 & \textbf{0.910} \\
Bronchiectasis & 0.781 & \textbf{0.800}\textsuperscript{$\dagger$} & 0.781 & 0.744 & 0.759 & 0.750 & 0.730 & 0.730 & 0.743 & 0.776 & 0.774 & \underline{0.792} & 0.787 & 0.723 & 0.740 & 0.685 & 0.786 \\
Interlobular septal thickening & 0.871 & \textbf{0.892}\textsuperscript{$\dagger$} & 0.871 & 0.848 & 0.840 & 0.847 & 0.803 & 0.820 & 0.842 & 0.862 & 0.868 & \underline{0.887} & \underline{0.887} & 0.835 & 0.846 & 0.823 & 0.873 \\
\midrule
\multicolumn{18}{l}{\textbf{(B) RAD-ChestCT held-out test}} \\
Arterial wall calcification & \textbf{0.717} & \underline{0.704} & 0.682\textsuperscript{$\dagger$} & 0.648 & 0.646 & 0.670 & 0.680 & 0.635 & 0.641 & 0.698 & 0.675 & 0.680 & 0.690 & 0.648 & 0.668 & 0.671 & 0.632 \\
Atelectasis & 0.655 & \textbf{0.758}\textsuperscript{$\dagger$} & 0.680 & 0.668 & 0.677 & 0.661 & 0.639 & 0.630 & 0.663 & 0.659 & 0.666 & 0.726 & 0.710 & 0.654 & 0.654 & 0.649 & \underline{0.733} \\
Bronchiectasis & 0.774 & \underline{0.796}\textsuperscript{$\dagger$} & 0.759 & 0.732 & \textbf{0.805} & 0.704 & 0.648 & 0.675 & 0.695 & 0.770 & 0.715 & 0.773 & 0.769 & 0.753 & 0.745 & 0.718 & 0.784 \\
Cardiomegaly & \textbf{0.946}\textsuperscript{$\dagger$} & 0.917 & 0.929 & 0.907 & 0.924 & 0.915 & 0.893 & 0.901 & 0.845 & 0.892 & 0.858 & \underline{0.943} & 0.923 & 0.901 & 0.903 & 0.893 & 0.822 \\
Consolidation & 0.709 & \underline{0.776}\textsuperscript{$\dagger$} & 0.726 & 0.688 & 0.700 & 0.695 & 0.714 & 0.630 & 0.704 & 0.704 & 0.748 & 0.729 & 0.737 & 0.678 & 0.682 & 0.687 & \textbf{0.814} \\
Coronary artery wall calcification & \textbf{0.787}\textsuperscript{$\dagger$} & 0.755 & \underline{0.780} & 0.750 & 0.744 & 0.772 & 0.746 & 0.731 & 0.712 & 0.749 & 0.762 & 0.766 & 0.768 & 0.707 & 0.751 & 0.779 & 0.709 \\
Emphysema & 0.857 & \textbf{0.909}\textsuperscript{$\dagger$} & 0.869 & 0.828 & 0.800 & 0.835 & 0.771 & 0.743 & 0.768 & 0.837 & 0.813 & 0.882 & \underline{0.890} & 0.819 & 0.805 & 0.729 & 0.836 \\
Hiatal hernia & 0.756 & 0.753 & 0.770 & 0.795 & 0.770 & 0.733 & 0.810 & 0.833 & 0.747 & \textbf{0.851}\textsuperscript{$\dagger$} & 0.799 & \underline{0.842} & 0.823 & 0.700 & 0.690 & 0.764 & 0.654 \\
Septal thickening & \underline{0.764} & \textbf{0.801}\textsuperscript{$\dagger$} & 0.726 & 0.693 & 0.653 & 0.664 & 0.618 & 0.614 & 0.537 & 0.678 & 0.674 & 0.701 & 0.701 & 0.678 & 0.644 & 0.685 & 0.752 \\
Nodule $>1$ cm & 0.585 & \underline{0.645}\textsuperscript{$\dagger$} & \textbf{0.653} & 0.611 & 0.586 & 0.568 & 0.516 & 0.576 & 0.549 & 0.635 & 0.533 & 0.524 & 0.594 & 0.581 & 0.550 & 0.550 & 0.614 \\
Opacity & 0.647 & \textbf{0.711}\textsuperscript{$\dagger$} & 0.676 & 0.603 & 0.613 & 0.630 & 0.622 & 0.641 & 0.591 & 0.677 & 0.661 & 0.661 & 0.663 & 0.602 & 0.620 & 0.676 & \underline{0.690} \\
Lymphadenopathy & 0.692 & 0.726 & 0.707 & 0.697 & 0.633 & 0.690 & 0.713 & 0.644 & 0.589 & \textbf{0.743} & \underline{0.741} & 0.710 & 0.723 & 0.683 & 0.685 & 0.673 & 0.678 \\
Bronchial wall thickening & 0.643 & \textbf{0.696}\textsuperscript{$\dagger$} & \underline{0.677} & 0.629 & 0.617 & 0.601 & 0.596 & 0.578 & 0.556 & 0.670 & 0.550 & 0.652 & 0.659 & 0.613 & 0.644 & 0.512 & 0.584 \\
Pericardial effusion & \textbf{0.791}\textsuperscript{$\dagger$} & 0.686 & 0.725 & 0.721 & 0.702 & 0.699 & 0.653 & 0.693 & 0.673 & 0.685 & 0.650 & \underline{0.749} & 0.718 & 0.677 & 0.670 & 0.714 & 0.665 \\
Pleural effusion & 0.921 & \underline{0.938}\textsuperscript{$\dagger$} & 0.914 & 0.911 & 0.918 & 0.897 & 0.885 & 0.867 & 0.893 & 0.898 & 0.872 & \textbf{0.939} & 0.924 & 0.853 & 0.887 & 0.902 & 0.922 \\
Fibrosis & 0.803 & \underline{0.835}\textsuperscript{$\dagger$} & 0.799 & 0.767 & 0.770 & 0.753 & 0.665 & 0.728 & 0.717 & 0.807 & 0.768 & 0.829 & 0.815 & 0.825 & 0.829 & 0.759 & \textbf{0.861} \\
Ground-glass opacity & 0.596 & \textbf{0.670}\textsuperscript{$\dagger$} & 0.597 & 0.555 & 0.533 & 0.569 & 0.537 & 0.486 & 0.513 & 0.552 & 0.525 & 0.585 & 0.574 & 0.552 & 0.597 & 0.549 & \underline{0.648} \\
Lung cancer & 0.769 & \underline{0.820}\textsuperscript{$\dagger$} & 0.771 & 0.716 & 0.750 & 0.767 & 0.675 & 0.659 & 0.721 & 0.753 & 0.724 & 0.786 & 0.786 & 0.687 & 0.724 & 0.655 & \textbf{0.841} \\
Lung infection & 0.557 & \textbf{0.622}\textsuperscript{$\dagger$} & 0.591 & 0.592 & 0.564 & 0.525 & \underline{0.621} & 0.599 & 0.554 & 0.585 & 0.584 & 0.612 & 0.598 & 0.531 & 0.582 & 0.593 & 0.608 \\
Lung scarring & 0.693 & \textbf{0.705}\textsuperscript{$\dagger$} & 0.694 & 0.666 & 0.680 & 0.690 & 0.637 & 0.662 & 0.649 & 0.642 & 0.655 & 0.696 & \underline{0.697} & 0.612 & 0.665 & 0.629 & 0.655 \\
Interstitial lung disease & 0.771 & \textbf{0.842}\textsuperscript{$\dagger$} & 0.781 & 0.755 & 0.736 & 0.747 & 0.675 & 0.709 & 0.693 & 0.737 & 0.753 & 0.794 & 0.789 & 0.773 & 0.783 & 0.765 & \underline{0.820} \\
Lung calcification & 0.631 & \textbf{0.647}\textsuperscript{$\dagger$} & 0.607 & 0.600 & 0.571 & 0.568 & 0.559 & 0.589 & \underline{0.634} & 0.571 & 0.569 & 0.591 & 0.610 & 0.581 & 0.571 & 0.633 & 0.608 \\
Aortic atherosclerosis & 0.642 & \textbf{0.660} & 0.640\textsuperscript{$\dagger$} & 0.607 & 0.635 & 0.624 & 0.640 & 0.567 & 0.579 & 0.627 & 0.575 & 0.621 & \underline{0.643} & 0.624 & 0.613 & 0.613 & 0.612 \\
\bottomrule
\end{tabular}}
\endgroup
\end{table*}

\section{Complete Prognostic Prediction Results}
\label{sec:complete-cox-results}
Supplementary Table~\ref{tab:cox-cindex} reports the complete test-set Harrell
C-indices for all organ-specific OrganLens representations, their concatenated
and mean aggregations, and the volume-level baselines on INSPECT and NLST.
Daggered OrganLens entries correspond to Figure~\ref{fig:main-results}.

\begin{table*}[t]
\centering
\caption{Complete prognostic prediction results. Panel A reports test-set Harrell C-indices on INSPECT, and Panel B reports test-set Harrell C-indices on NLST. Best and second-best values within each endpoint are shown in bold and underlined, respectively. Ties after rounding to three decimal places receive the same formatting. A dagger marks the anatomically matched OrganLens representation used in Figure~\ref{fig:main-results}. Concat and mean aggregate all 11 organ-specific representations. SPECTRE was trained on INSPECT and NLST. An asterisk marks the CT-pretrained DINOv2 Cox model for NLST IHD mortality, for which two additional random seeds gave 0.602 and 0.604.}
\label{tab:cox-cindex}
\begingroup
\renewcommand{\arraystretch}{0.96}
\resizebox{\textwidth}{!}{%
\begin{tabular}{lcccccccc}
\toprule
Representation & Mortality & Readmission & Pulmonary Hypertension & Pleural Effusion & Consolidation & Edema & Atelectasis & Cardiomegaly \\
\midrule
\multicolumn{9}{l}{\textbf{(A) INSPECT}} \\
\multicolumn{9}{l}{\textbf{OrganLens organ-specific representations}} \\
\quad Heart & 0.783 & 0.593 & \textbf{0.742}\textsuperscript{$\dagger$} & 0.818 & 0.749 & 0.661 & 0.738 & \textbf{0.745}\textsuperscript{$\dagger$} \\
\quad Spleen & 0.768 & 0.586 & 0.702 & 0.801 & 0.713 & 0.639 & 0.714 & 0.703 \\
\quad Kidneys & 0.774 & 0.589 & 0.695 & 0.807 & 0.734 & 0.655 & 0.727 & 0.704 \\
\quad Liver & 0.767 & 0.607 & 0.687 & 0.818 & 0.705 & 0.654 & 0.722 & 0.692 \\
\quad Pancreas & 0.783 & 0.601 & 0.702 & 0.819 & 0.757 & 0.663 & 0.741 & 0.711 \\
\quad Lung & \textbf{0.795} & 0.591 & \underline{0.736} & \textbf{0.836}\textsuperscript{$\dagger$} & \textbf{0.778}\textsuperscript{$\dagger$} & \textbf{0.665}\textsuperscript{$\dagger$} & \textbf{0.756}\textsuperscript{$\dagger$} & 0.735 \\
\quad Intestine & 0.764 & 0.578 & 0.694 & 0.795 & 0.719 & 0.645 & 0.699 & 0.693 \\
\quad Aorta & 0.784 & 0.597 & 0.700 & 0.813 & 0.756 & 0.663 & 0.743 & 0.716 \\
\quad Stomach & 0.765 & 0.572 & 0.689 & 0.794 & 0.697 & 0.651 & 0.703 & 0.701 \\
\quad Esophagus & 0.786 & 0.598 & 0.701 & 0.819 & 0.759 & \underline{0.664} & 0.744 & 0.699 \\
\quad Trachea & 0.786 & 0.597 & 0.697 & 0.819 & 0.760 & \underline{0.664} & 0.745 & 0.696 \\
\multicolumn{9}{l}{\textbf{OrganLens global aggregations}} \\
\quad Concat & \underline{0.794}\textsuperscript{$\dagger$} & \underline{0.608}\textsuperscript{$\dagger$} & 0.732 & \underline{0.832} & 0.775 & 0.656 & \underline{0.751} & \underline{0.736} \\
\quad Mean & 0.784 & 0.590 & 0.711 & 0.822 & 0.753 & 0.663 & 0.738 & 0.715 \\
\midrule
\multicolumn{9}{l}{\textbf{Volume-level baselines}} \\
\quad CT-pretrained DINOv2 & 0.784 & 0.592 & 0.692 & 0.802 & 0.746 & 0.659 & 0.741 & 0.709 \\
\quad GigaHeart & 0.782 & 0.602 & 0.709 & 0.811 & 0.741 & 0.642 & 0.735 & 0.706 \\
\quad Merlin & 0.759 & 0.586 & 0.676 & 0.806 & 0.718 & 0.654 & 0.708 & 0.671 \\
\quad SPECTRE & 0.767 & \textbf{0.611} & 0.717 & 0.821 & \underline{0.777} & 0.635 & 0.736 & 0.704 \\
\bottomrule
\end{tabular}}
\par\vspace{5pt}
\resizebox{\textwidth}{!}{%
\begin{tabular}{lccccccccc}
\toprule
Representation & LC inc. & All-cause & LC death & CV & IHD & Resp. & COPD & Any cancer & Non-LC \\
\midrule
\multicolumn{10}{l}{\textbf{(B) NLST}} \\
\multicolumn{10}{l}{\textbf{OrganLens organ-specific representations}} \\
\quad Heart & 0.648 & 0.732 & 0.686 & \textbf{0.753}\textsuperscript{$\dagger$} & \underline{0.752}\textsuperscript{$\dagger$} & 0.886 & 0.935 & 0.663 & \underline{0.738} \\
\quad Spleen & 0.663 & 0.716 & 0.713 & 0.662 & 0.625 & 0.840 & 0.817 & 0.667 & 0.719 \\
\quad Kidneys & 0.671 & 0.726 & 0.723 & 0.683 & 0.678 & 0.892 & 0.929 & 0.667 & 0.705 \\
\quad Liver & 0.635 & 0.713 & 0.684 & 0.675 & 0.638 & 0.851 & 0.863 & 0.668 & 0.727 \\
\quad Pancreas & 0.682 & 0.737 & 0.737 & 0.716 & 0.723 & 0.910 & 0.947 & 0.681 & 0.733 \\
\quad Lung & \underline{0.709}\textsuperscript{$\dagger$} & 0.748 & \textbf{0.780}\textsuperscript{$\dagger$} & 0.708 & 0.743 & \textbf{0.935}\textsuperscript{$\dagger$} & \textbf{0.960}\textsuperscript{$\dagger$} & 0.696 & 0.731 \\
\quad Intestine & 0.652 & 0.711 & 0.691 & 0.691 & 0.665 & 0.864 & 0.879 & 0.643 & 0.703 \\
\quad Aorta & 0.681 & 0.732 & 0.735 & 0.726 & 0.726 & 0.898 & 0.934 & 0.675 & 0.735 \\
\quad Stomach & 0.662 & 0.705 & 0.705 & 0.705 & 0.710 & 0.824 & 0.813 & 0.664 & 0.713 \\
\quad Esophagus & 0.686 & 0.738 & 0.746 & 0.718 & 0.731 & 0.908 & 0.949 & 0.687 & 0.737 \\
\quad Trachea & 0.683 & 0.737 & 0.741 & 0.717 & 0.727 & 0.908 & 0.950 & 0.684 & 0.735 \\
\multicolumn{10}{l}{\textbf{OrganLens global aggregations}} \\
\quad Concat & 0.683 & \textbf{0.748}\textsuperscript{$\dagger$} & 0.755 & 0.741 & 0.749 & \underline{0.928} & \underline{0.951} & \textbf{0.707}\textsuperscript{$\dagger$} & \textbf{0.756}\textsuperscript{$\dagger$} \\
\quad Mean & 0.686 & \underline{0.740} & 0.742 & 0.734 & 0.744 & 0.901 & 0.935 & \underline{0.698} & 0.732 \\
\midrule
\multicolumn{10}{l}{\textbf{Volume-level baselines}} \\
\quad CT-pretrained DINOv2 & 0.640 & 0.701 & 0.683 & 0.638 & 0.488\textsuperscript{*} & 0.860 & 0.898 & 0.597 & 0.674 \\
\quad GigaHeart & 0.647 & 0.723 & 0.695 & 0.672 & 0.695 & 0.886 & 0.911 & 0.632 & 0.697 \\
\quad Merlin & 0.639 & 0.697 & 0.676 & \underline{0.747} & 0.734 & 0.860 & 0.873 & 0.640 & 0.713 \\
\quad SPECTRE & \textbf{0.778} & 0.721 & \underline{0.777} & 0.737 & \textbf{0.766} & 0.900 & 0.930 & 0.630 & 0.705 \\
\bottomrule
\end{tabular}}
\endgroup
\end{table*}

\section{Longitudinal Baseline Comparison}
\label{sec:longitudinal-baseline-comparison}

\begin{figure*}[t]
  \centering
  \includegraphics[width=\textwidth]{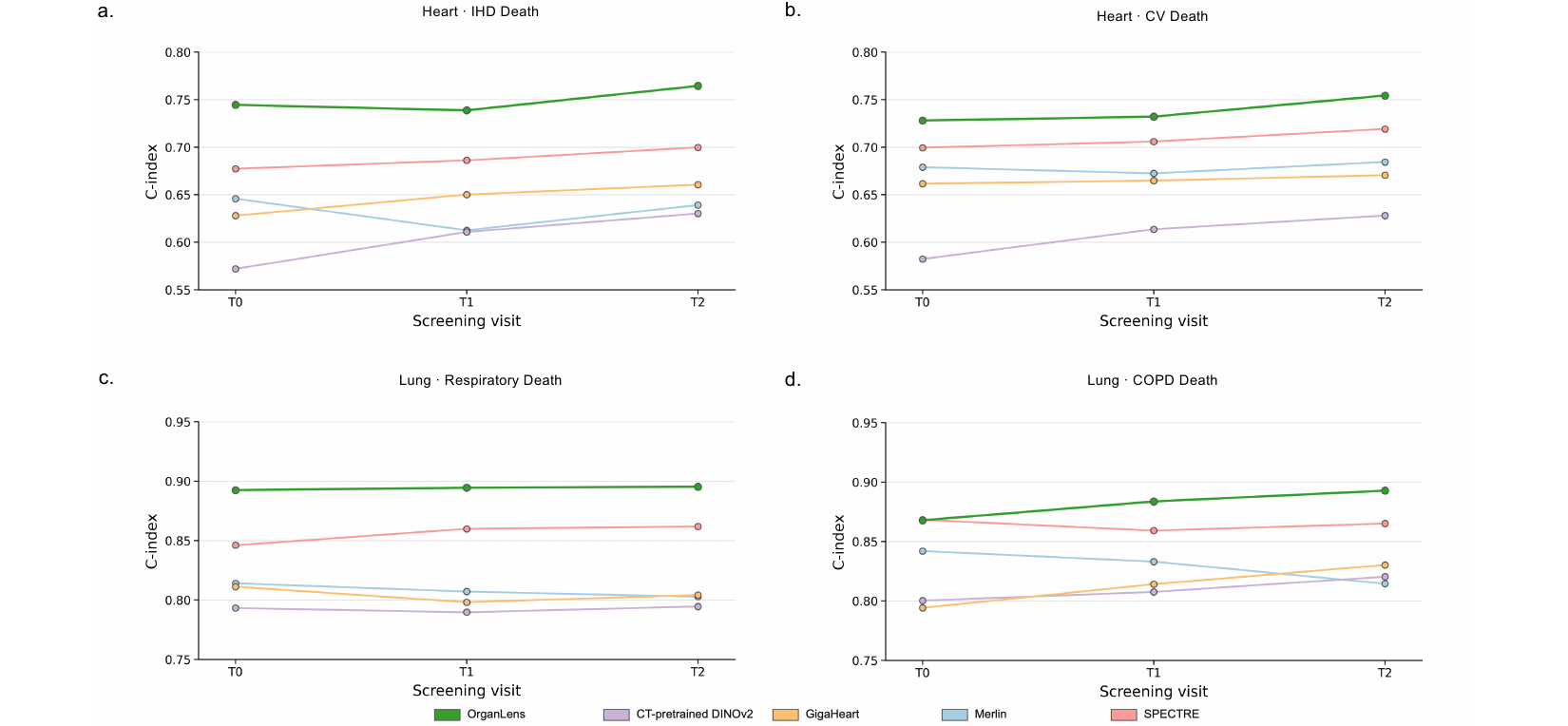}
  \caption{Longitudinal prognostic comparison on NLST. Visit-specific Harrell
  C-indices at T0--T2 are shown for (a) ischemic-heart-disease mortality,
  (b) cardiovascular mortality, (c) respiratory mortality, and (d) COPD
  mortality. OrganLens uses the heart representation for cardiovascular
  endpoints and the lung representation for respiratory endpoints. All
  methods are evaluated on the common longitudinal test cohort.}
  \Description{Four line plots compare the visit-specific Harrell C-indices
  of OrganLens, CT-pretrained DINOv2, GigaHeart, Merlin, and SPECTRE across
  T0, T1, and T2 for two cardiovascular and two respiratory mortality
  endpoints.}
  \label{fig:longitudinal-baselines}
\end{figure*}

Supplementary Figure~\ref{fig:longitudinal-baselines} compares visit-specific
Harrell C-indices across the common NLST longitudinal test cohort. OrganLens
used the heart representation for cardiovascular endpoints and the lung
representation for respiratory endpoints. It produced the highest point
estimates at T0--T2 for cardiovascular, ischemic-heart-disease, and respiratory
mortality. For COPD mortality, OrganLens and SPECTRE were comparable at T0,
while OrganLens was higher at T1 and T2.

\begin{figure}[t]
  \centering
  \includegraphics[width=\columnwidth]{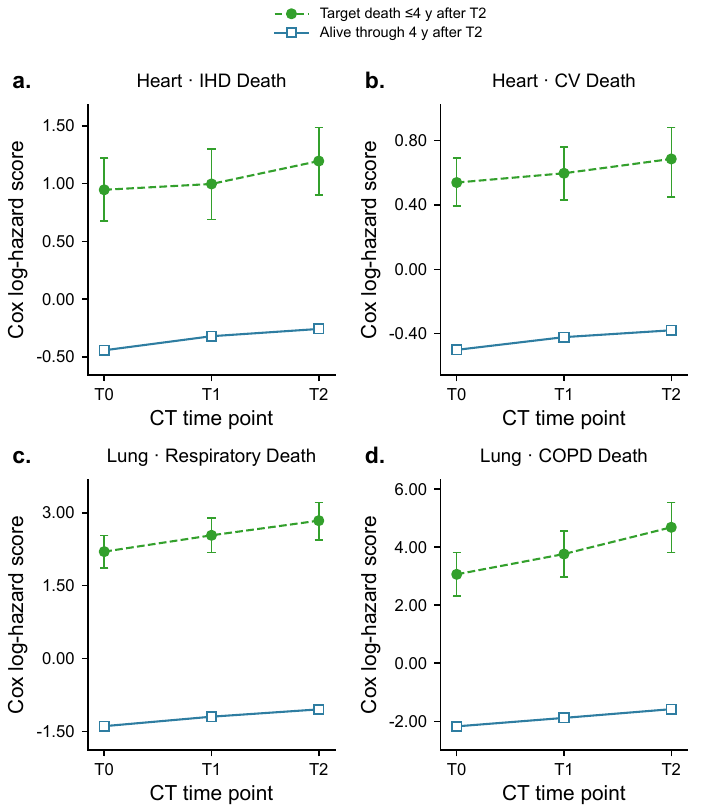}
  \caption{Longitudinal NLST scores with survivor controls. Mean anatomically
  matched Cox log-hazard scores at T0, T1, and T2 for participants who died of
  the target cause within four years after T2 and for participants alive
  through four years after T2 ($n=19{,}341$): (a) ischemic-heart-disease
  mortality ($n=137$), (b) cardiovascular mortality ($n=245$), (c) respiratory
  mortality ($n=103$), and (d) COPD mortality ($n=71$). Error bars are 95\%
  confidence intervals from 2,000 participant-level bootstrap resamples.}
  \Description{Four line plots show mean Cox log-hazard scores at T0, T1, and
  T2 for participants who died of the target cause within four years after T2
  and for survivors, for ischemic-heart-disease, cardiovascular, respiratory,
  and COPD mortality.}
  \label{fig:longitudinal-controls}
\end{figure}
\section{Complete Image--Text Retrieval Results}
\label{sec:complete-retrieval-results}
Supplementary Table~\ref{tab:complete-retrieval-results} reports Recall@$K$ in both
retrieval directions for every evaluated model. Within each cohort, all
methods use the same fixed candidate set. For OrganLens, both concatenated and
mean aggregations of the 11 organ-specific representations are reported. The
OrganLens representations are computed from 64 uniformly sampled axial slices.

\begin{table*}[t]
\centering
\caption{Complete bidirectional image--text retrieval results. Values are
Recall@$K$ (\%). Best and second-best values within each cohort and direction
are shown in bold and underlined, respectively. Ties after rounding to two
decimal places receive the same formatting. SPECTRE was trained on the CT-RATE
training split and INSPECT. The main OrganLens representation concatenates all
11 organ-specific representations. Mean aggregation averages them instead.}
\label{tab:complete-retrieval-results}
\begin{tabular*}{\textwidth}{@{\extracolsep{\fill}}lcccccccc}
\toprule
& \multicolumn{4}{c}{Text-to-image} & \multicolumn{4}{c}{Image-to-text} \\
\cmidrule(lr){2-5}\cmidrule(lr){6-9}
Method & R@5 & R@10 & R@50 & R@100 & R@5 & R@10 & R@50 & R@100 \\
\midrule
\multicolumn{9}{@{}l}{\textbf{CT-RATE}} \\
\addlinespace[1pt]
\multicolumn{9}{@{}l}{\textit{OrganLens}} \\
\quad Concat & \underline{36.45} & \underline{46.55} & \textbf{69.18} & \underline{78.52} & \textbf{37.08} & \textbf{46.42} & \underline{69.05} & \textbf{79.09} \\
\quad Mean & \textbf{37.47} & \textbf{46.93} & \underline{68.93} & \textbf{79.41} & \underline{36.25} & \underline{45.72} & \textbf{69.18} & \underline{78.45} \\
\addlinespace[2pt]
CT-pretrained DINOv2 & 27.88 & 36.32 & 62.66 & 74.49 & 26.21 & 36.32 & 62.28 & 75.38 \\
GigaHeart & 28.58 & 37.79 & 66.43 & 78.32 & 26.85 & 36.38 & 64.83 & 76.85 \\
Merlin & 21.93 & 30.12 & 58.31 & 70.08 & 22.38 & 31.39 & 56.78 & 69.88 \\
CT-CLIP & 5.50 & 8.95 & 27.75 & 42.52 & 4.54 & 8.89 & 26.41 & 41.88 \\
SPECTRE & 22.44 & 31.52 & 56.59 & 67.33 & 22.12 & 31.39 & 57.74 & 67.97 \\
\midrule
\multicolumn{9}{@{}l}{\textbf{INSPECT}} \\
\addlinespace[1pt]
\multicolumn{9}{@{}l}{\textit{OrganLens}} \\
\quad Concat & \textbf{23.88} & \textbf{33.09} & \textbf{58.37} & \textbf{68.90} & \textbf{23.19} & \textbf{32.04} & \textbf{57.25} & \textbf{68.43} \\
\quad Mean & \underline{22.63} & \underline{31.66} & \underline{57.35} & \underline{68.46} & \underline{22.73} & \underline{31.29} & \underline{56.76} & \underline{68.06} \\
\addlinespace[2pt]
CT-pretrained DINOv2 & 15.29 & 22.38 & 47.60 & 61.08 & 13.61 & 21.26 & 45.14 & 58.00 \\
GigaHeart & 11.61 & 18.15 & 41.81 & 55.32 & 11.74 & 17.81 & 40.78 & 53.89 \\
Merlin & 10.37 & 16.16 & 39.79 & 53.77 & 9.71 & 15.57 & 38.89 & 52.77 \\
CT-CLIP & 4.92 & 8.75 & 25.37 & 38.45 & 6.01 & 10.18 & 26.87 & 38.67 \\
SPECTRE & 15.47 & 23.85 & 51.77 & 67.12 & 15.82 & 23.57 & 51.62 & 66.44 \\
\bottomrule
\end{tabular*}
\end{table*}

\section{Complete Ablation Results}
\label{sec:complete-ablation}
Supplementary Table~\ref{tab:pretraining-ablation} compares the two-by-two
combinations of organ-guided cropping and anatomy-mask supervision, and
Supplementary Table~\ref{tab:pooling-ablation} compares four pooling
strategies. Both report the 16 CT-RATE and 22 RAD-ChestCT abnormalities with
predefined anatomical matches, including the matched organ-specific
representation and the mean and concatenated aggregations of all 11
organ-specific representations.

\begin{table*}[t]
\centering
\caption{Complete pretraining-component ablation results. Panels A and B report AUROCs on the CT-RATE validation split and RAD-ChestCT held-out test split for the 16 and 22 abnormalities with predefined anatomical matches, respectively. Matched uses the organ-specific representation assigned to each abnormality, Mean averages the 11 organ-specific representations, and Concat concatenates them. The variants form a two-by-two comparison of organ-guided cropping and anatomy-mask supervision. Variants without anatomy-mask supervision use mean CLS pooling; variants with it use anatomy-weighted pooling. The best and second-best values within each abnormality are shown in bold and underlined, respectively. Each style marks one value; ties after rounding to three decimal places favor OrganLens (Full) when applicable.}
\label{tab:pretraining-ablation}
\begingroup
\setlength{\tabcolsep}{2.5pt}
\renewcommand{\arraystretch}{0.96}
\resizebox{\textwidth}{!}{%
\begin{tabular}{l*{12}{c}}
\toprule
& \multicolumn{3}{c}{\shortstack{No organ-guided crops\\or anatomy-mask supervision}}
& \multicolumn{3}{c}{\shortstack{No anatomy-mask\\supervision}}
& \multicolumn{3}{c}{\shortstack{No organ-guided\\crops}}
& \multicolumn{3}{c}{OrganLens (Full)} \\
\cmidrule(lr){2-4}\cmidrule(lr){5-7}\cmidrule(lr){8-10}\cmidrule(lr){11-13}
Abnormality
& Matched & Mean & Concat
& Matched & Mean & Concat
& Matched & Mean & Concat
& Matched & Mean & Concat \\
\midrule
\multicolumn{13}{l}{\textbf{(A) CT-RATE validation}} \\
Arterial wall calcification & 0.892 & 0.888 & 0.901 & 0.896 & 0.895 & 0.906 & \textbf{0.929} & 0.920 & 0.923 & \underline{0.928} & 0.922 & 0.921 \\
Cardiomegaly & 0.926 & 0.925 & 0.917 & 0.940 & 0.939 & 0.933 & \textbf{0.954} & 0.950 & 0.950 & \underline{0.953} & 0.949 & 0.950 \\
Pericardial effusion & 0.813 & 0.797 & 0.822 & 0.834 & 0.842 & 0.852 & \underline{0.920} & 0.882 & 0.907 & \textbf{0.925} & 0.897 & 0.910 \\
Coronary artery wall calcification & 0.869 & 0.868 & 0.874 & 0.876 & 0.876 & 0.890 & 0.903 & 0.905 & 0.909 & \underline{0.905} & \textbf{0.909} & 0.905 \\
Hiatal hernia & 0.728 & 0.744 & 0.746 & 0.754 & 0.747 & 0.750 & 0.845 & 0.820 & 0.838 & \textbf{0.849} & 0.833 & \underline{0.846} \\
Emphysema & 0.760 & 0.751 & 0.773 & 0.766 & 0.765 & 0.792 & \underline{0.818} & 0.806 & 0.803 & \textbf{0.819} & 0.807 & 0.804 \\
Atelectasis & 0.718 & 0.715 & 0.728 & 0.725 & 0.723 & 0.743 & \underline{0.777} & 0.761 & 0.768 & \textbf{0.782} & 0.763 & 0.765 \\
Lung nodule & 0.642 & 0.650 & 0.662 & 0.652 & 0.653 & 0.672 & 0.696 & 0.696 & 0.692 & \textbf{0.696} & \underline{0.694} & 0.647 \\
Lung opacity & 0.779 & 0.767 & 0.783 & 0.791 & 0.792 & 0.790 & \underline{0.832} & 0.799 & 0.814 & \textbf{0.839} & 0.816 & 0.824 \\
Pulmonary fibrotic sequela & 0.651 & 0.655 & 0.660 & 0.658 & 0.659 & 0.672 & \underline{0.712} & 0.691 & 0.699 & \textbf{0.714} & 0.694 & 0.700 \\
Pleural effusion & 0.962 & 0.963 & 0.961 & 0.965 & 0.963 & 0.966 & 0.971 & 0.969 & 0.971 & \textbf{0.974} & 0.967 & \underline{0.972} \\
Mosaic attenuation pattern & 0.847 & 0.839 & 0.861 & 0.874 & 0.870 & 0.883 & \underline{0.903} & 0.884 & 0.892 & \textbf{0.906} & 0.885 & 0.897 \\
Peribronchial thickening & 0.770 & 0.751 & 0.796 & 0.773 & 0.769 & 0.802 & 0.825 & 0.812 & 0.814 & \underline{0.823} & \textbf{0.825} & 0.825 \\
Consolidation & 0.824 & 0.821 & 0.828 & 0.835 & 0.834 & 0.839 & \textbf{0.888} & 0.843 & 0.854 & \underline{0.887} & 0.857 & 0.854 \\
Bronchiectasis & 0.738 & 0.720 & 0.742 & 0.744 & 0.741 & 0.763 & 0.788 & 0.776 & 0.782 & \textbf{0.800} & 0.787 & \underline{0.792} \\
Interlobular septal thickening & 0.843 & 0.842 & 0.859 & 0.858 & 0.852 & 0.861 & \underline{0.889} & 0.876 & 0.879 & \textbf{0.892} & 0.887 & 0.887 \\
\midrule
\multicolumn{13}{l}{\textbf{(B) RAD-ChestCT held-out test}} \\
Arterial wall calcification & 0.691 & 0.686 & 0.676 & 0.690 & \underline{0.693} & \textbf{0.702} & 0.678 & 0.666 & 0.677 & 0.682 & 0.690 & 0.680 \\
Atelectasis & 0.708 & 0.707 & 0.707 & 0.720 & 0.716 & 0.719 & \underline{0.755} & 0.708 & 0.695 & \textbf{0.758} & 0.710 & 0.726 \\
Bronchiectasis & 0.766 & 0.768 & 0.755 & 0.779 & 0.781 & 0.781 & \textbf{0.804} & 0.785 & 0.766 & \underline{0.796} & 0.769 & 0.773 \\
Cardiomegaly & 0.922 & 0.918 & 0.925 & 0.935 & 0.930 & 0.926 & \textbf{0.947} & 0.932 & 0.939 & \underline{0.946} & 0.923 & 0.943 \\
Consolidation & 0.745 & 0.736 & 0.722 & 0.745 & 0.743 & 0.731 & \underline{0.764} & 0.747 & 0.739 & \textbf{0.776} & 0.737 & 0.729 \\
Coronary artery wall calcification & 0.743 & 0.742 & 0.760 & 0.779 & 0.778 & 0.777 & \underline{0.786} & 0.764 & 0.758 & \textbf{0.787} & 0.768 & 0.766 \\
Emphysema & 0.861 & 0.861 & 0.826 & 0.821 & 0.837 & 0.837 & \textbf{0.910} & 0.876 & 0.878 & \underline{0.909} & 0.890 & 0.882 \\
Hiatal hernia & 0.704 & 0.699 & 0.676 & 0.756 & 0.736 & 0.756 & 0.838 & 0.774 & 0.828 & \textbf{0.851} & 0.823 & \underline{0.842} \\
Septal thickening & 0.668 & 0.670 & 0.635 & 0.722 & 0.722 & 0.682 & \textbf{0.809} & 0.721 & 0.680 & \underline{0.801} & 0.701 & 0.701 \\
Nodule $>1$ cm & 0.545 & 0.539 & 0.582 & 0.575 & 0.538 & 0.568 & 0.614 & 0.600 & \underline{0.625} & \textbf{0.645} & 0.594 & 0.524 \\
Opacity & 0.629 & 0.638 & 0.644 & 0.670 & 0.674 & 0.651 & \textbf{0.724} & 0.636 & 0.637 & \underline{0.711} & 0.663 & 0.661 \\
Bronchial wall thickening & 0.599 & 0.598 & 0.600 & 0.625 & 0.626 & 0.686 & 0.669 & \textbf{0.733} & 0.649 & \underline{0.696} & 0.659 & 0.652 \\
Pericardial effusion & 0.689 & 0.692 & 0.692 & 0.702 & 0.685 & 0.698 & \textbf{0.796} & 0.738 & 0.738 & \underline{0.791} & 0.718 & 0.749 \\
Pleural effusion & 0.911 & 0.911 & 0.912 & 0.921 & 0.928 & 0.916 & 0.936 & 0.928 & 0.927 & \underline{0.938} & 0.924 & \textbf{0.939} \\
Fibrosis & 0.850 & \underline{0.852} & 0.813 & 0.844 & 0.844 & 0.839 & 0.848 & \textbf{0.857} & 0.817 & 0.835 & 0.815 & 0.829 \\
Ground-glass opacity & 0.580 & 0.581 & 0.557 & 0.596 & 0.596 & 0.568 & \underline{0.650} & 0.548 & 0.556 & \textbf{0.670} & 0.574 & 0.585 \\
Lung cancer & 0.764 & 0.767 & 0.764 & 0.774 & 0.771 & 0.758 & \textbf{0.823} & 0.765 & 0.772 & \underline{0.820} & 0.786 & 0.786 \\
Lung infection & 0.545 & 0.592 & 0.567 & 0.568 & 0.570 & 0.548 & 0.612 & \textbf{0.627} & 0.577 & \underline{0.622} & 0.598 & 0.612 \\
Lung scarring & 0.666 & 0.642 & 0.632 & 0.680 & \textbf{0.711} & 0.678 & 0.696 & 0.695 & 0.678 & \underline{0.705} & 0.697 & 0.696 \\
Interstitial lung disease & 0.818 & 0.818 & 0.813 & 0.815 & 0.818 & 0.814 & \underline{0.836} & 0.802 & 0.773 & \textbf{0.842} & 0.789 & 0.794 \\
Lung calcification & 0.601 & 0.601 & 0.594 & 0.597 & 0.599 & 0.598 & \underline{0.620} & 0.592 & 0.527 & \textbf{0.647} & 0.610 & 0.591 \\
Aortic atherosclerosis & 0.630 & 0.621 & 0.584 & 0.592 & 0.591 & 0.579 & 0.624 & \textbf{0.645} & 0.631 & 0.640 & \underline{0.643} & 0.621 \\
\bottomrule
\end{tabular}}
\endgroup
\end{table*}
\begin{table*}[t]
\centering
\caption{Complete pooling ablation results. Panels A and B report AUROCs on the CT-RATE validation split and RAD-ChestCT held-out test split for the 16 and 22 abnormalities with predefined anatomical matches, respectively. Matched uses the organ-specific representation assigned to each abnormality, Mean averages the 11 organ-specific representations, and Concat concatenates them. Mean CLS Pooling averages organ-conditioned slice CLS features. Uniform Patch Pooling averages all patch features. The w/o Inter-slice Weighting variant weights patches by predicted spatial support but averages slices uniformly. Full additionally weights slices by predicted organ area. All variants retain all sampled slices. The Full group reproduces the complete-method results in Supplementary Table~\ref{tab:abnormality-organ-auroc}. The best and second-best values within each abnormality are shown in bold and underlined, respectively. Each style marks one value. Ties after rounding to three decimal places favor OrganLens (Full) when applicable.}
\label{tab:pooling-ablation}
\begingroup
\setlength{\tabcolsep}{2.5pt}
\renewcommand{\arraystretch}{0.96}
\resizebox{\textwidth}{!}{%
\begin{tabular}{l*{12}{c}}
\toprule
& \multicolumn{3}{c}{\shortstack{OrganLens w/\\Mean CLS Pooling}}
& \multicolumn{3}{c}{\shortstack{OrganLens w/\\Uniform Patch Pooling}}
& \multicolumn{3}{c}{\shortstack{OrganLens w/o\\Inter-slice Weighting}}
& \multicolumn{3}{c}{OrganLens (Full)} \\
\cmidrule(lr){2-4}\cmidrule(lr){5-7}\cmidrule(lr){8-10}\cmidrule(lr){11-13}
Abnormality
& Matched & Mean & Concat
& Matched & Mean & Concat
& Matched & Mean & Concat
& Matched & Mean & Concat \\
\midrule
\multicolumn{13}{l}{\textbf{(A) CT-RATE validation}} \\
Arterial wall calcification & 0.895 & 0.894 & 0.898 & 0.914 & 0.913 & 0.913 & \underline{0.925} & 0.923 & 0.921 & \textbf{0.928} & 0.922 & 0.921 \\
Cardiomegaly & 0.924 & 0.927 & 0.921 & 0.940 & 0.934 & 0.935 & 0.949 & 0.947 & 0.948 & \textbf{0.953} & 0.949 & \underline{0.950} \\
Pericardial effusion & 0.813 & 0.804 & 0.830 & 0.861 & 0.854 & 0.857 & 0.906 & 0.892 & 0.894 & \textbf{0.925} & 0.897 & \underline{0.910} \\
Coronary artery wall calcification & 0.877 & 0.875 & 0.880 & 0.901 & 0.899 & 0.891 & \underline{0.908} & 0.908 & 0.908 & 0.905 & \textbf{0.909} & 0.905 \\
Hiatal hernia & 0.743 & 0.736 & 0.740 & 0.763 & 0.762 & 0.767 & 0.836 & 0.826 & 0.836 & \textbf{0.849} & 0.833 & \underline{0.846} \\
Emphysema & 0.771 & 0.781 & 0.787 & 0.815 & 0.808 & 0.806 & \textbf{0.822} & 0.809 & 0.804 & \underline{0.819} & 0.807 & 0.804 \\
Atelectasis & 0.719 & 0.729 & 0.726 & 0.748 & 0.739 & 0.735 & \underline{0.769} & 0.760 & 0.763 & \textbf{0.782} & 0.763 & 0.765 \\
Lung nodule & 0.640 & 0.646 & 0.638 & 0.691 & 0.683 & 0.572 & 0.696 & 0.694 & 0.692 & \textbf{0.696} & \underline{0.694} & 0.647 \\
Lung opacity & 0.759 & 0.760 & 0.775 & 0.814 & 0.788 & 0.784 & \underline{0.834} & 0.816 & 0.825 & \textbf{0.839} & 0.816 & 0.824 \\
Pulmonary fibrotic sequela & 0.675 & 0.666 & 0.677 & 0.696 & 0.690 & 0.691 & \underline{0.709} & 0.700 & 0.703 & \textbf{0.714} & 0.694 & 0.700 \\
Pleural effusion & 0.958 & 0.960 & 0.959 & 0.964 & 0.960 & 0.960 & 0.972 & 0.964 & 0.969 & \textbf{0.974} & 0.967 & \underline{0.972} \\
Mosaic attenuation pattern & 0.853 & 0.868 & 0.875 & 0.879 & 0.876 & 0.876 & \underline{0.900} & 0.886 & 0.892 & \textbf{0.906} & 0.885 & 0.897 \\
Peribronchial thickening & 0.770 & 0.762 & 0.775 & 0.811 & 0.804 & 0.807 & 0.823 & 0.820 & 0.822 & \underline{0.823} & \textbf{0.825} & 0.825 \\
Consolidation & 0.820 & 0.821 & 0.832 & 0.839 & 0.812 & 0.807 & \underline{0.878} & 0.861 & 0.861 & \textbf{0.887} & 0.857 & 0.854 \\
Bronchiectasis & 0.747 & 0.752 & 0.770 & 0.790 & 0.783 & 0.783 & \underline{0.799} & 0.794 & 0.796 & \textbf{0.800} & 0.787 & 0.792 \\
Interlobular septal thickening & 0.864 & 0.856 & 0.857 & 0.880 & 0.865 & 0.873 & \underline{0.888} & 0.883 & 0.883 & \textbf{0.892} & 0.887 & 0.887 \\
\midrule
\multicolumn{13}{l}{\textbf{(B) RAD-ChestCT held-out test}} \\
Arterial wall calcification & 0.647 & 0.641 & 0.639 & 0.682 & 0.685 & \textbf{0.698} & \underline{0.695} & 0.688 & 0.695 & 0.682 & 0.690 & 0.680 \\
Atelectasis & 0.685 & 0.696 & 0.723 & 0.695 & 0.679 & 0.673 & \underline{0.735} & 0.703 & 0.706 & \textbf{0.758} & 0.710 & 0.726 \\
Bronchiectasis & \textbf{0.800} & 0.793 & 0.775 & 0.765 & 0.747 & 0.728 & \underline{0.797} & 0.767 & 0.777 & 0.796 & 0.769 & 0.773 \\
Cardiomegaly & 0.917 & 0.894 & 0.879 & 0.933 & 0.906 & 0.910 & 0.940 & 0.920 & 0.924 & \textbf{0.946} & 0.923 & \underline{0.943} \\
Consolidation & 0.729 & \underline{0.749} & 0.700 & 0.698 & 0.692 & 0.695 & 0.746 & 0.725 & 0.722 & \textbf{0.776} & 0.737 & 0.729 \\
Coronary artery wall calcification & 0.766 & 0.766 & 0.745 & 0.778 & 0.772 & 0.770 & \underline{0.783} & 0.776 & 0.769 & \textbf{0.787} & 0.768 & 0.766 \\
Emphysema & 0.838 & 0.834 & 0.842 & 0.882 & 0.875 & 0.880 & \underline{0.908} & 0.864 & 0.888 & \textbf{0.909} & 0.890 & 0.882 \\
Hiatal hernia & 0.745 & 0.721 & 0.734 & 0.764 & 0.753 & 0.620 & 0.835 & 0.789 & 0.822 & \textbf{0.851} & 0.823 & \underline{0.842} \\
Septal thickening & 0.711 & 0.686 & 0.651 & 0.711 & 0.664 & 0.659 & \underline{0.757} & 0.708 & 0.715 & \textbf{0.801} & 0.701 & 0.701 \\
Nodule $>1$ cm & 0.546 & 0.555 & 0.546 & 0.602 & 0.550 & 0.447 & 0.589 & \underline{0.609} & 0.550 & \textbf{0.645} & 0.594 & 0.524 \\
Opacity & 0.624 & 0.656 & 0.645 & 0.629 & 0.646 & 0.641 & \underline{0.681} & 0.648 & 0.671 & \textbf{0.711} & 0.663 & 0.661 \\
Bronchial wall thickening & 0.659 & 0.644 & 0.654 & 0.606 & 0.544 & 0.536 & 0.659 & 0.631 & 0.635 & \textbf{0.696} & \underline{0.659} & 0.652 \\
Pericardial effusion & 0.690 & 0.656 & 0.669 & 0.725 & 0.709 & 0.724 & \underline{0.766} & 0.752 & 0.745 & \textbf{0.791} & 0.718 & 0.749 \\
Pleural effusion & 0.912 & 0.909 & 0.903 & 0.935 & 0.931 & 0.931 & \textbf{0.942} & 0.934 & 0.938 & 0.938 & 0.924 & \underline{0.939} \\
Fibrosis & 0.819 & 0.820 & \textbf{0.855} & 0.824 & 0.825 & 0.826 & 0.828 & 0.813 & 0.811 & \underline{0.835} & 0.815 & 0.829 \\
Ground-glass opacity & 0.580 & 0.575 & 0.548 & 0.593 & 0.605 & 0.580 & \underline{0.638} & 0.558 & 0.569 & \textbf{0.670} & 0.574 & 0.585 \\
Lung cancer & 0.745 & 0.758 & 0.759 & 0.791 & 0.788 & 0.790 & \underline{0.805} & 0.787 & 0.791 & \textbf{0.820} & 0.786 & 0.786 \\
Lung infection & 0.584 & 0.600 & 0.580 & 0.557 & 0.538 & 0.521 & 0.611 & 0.618 & \textbf{0.625} & \underline{0.622} & 0.598 & 0.612 \\
Lung scarring & 0.652 & 0.665 & 0.653 & 0.682 & 0.669 & 0.667 & 0.687 & 0.695 & \underline{0.698} & \textbf{0.705} & 0.697 & 0.696 \\
Interstitial lung disease & 0.830 & \underline{0.837} & 0.823 & 0.804 & 0.805 & 0.797 & 0.816 & 0.800 & 0.776 & \textbf{0.842} & 0.789 & 0.794 \\
Lung calcification & 0.567 & 0.598 & 0.593 & 0.607 & 0.599 & \underline{0.614} & 0.597 & 0.597 & 0.603 & \textbf{0.647} & 0.610 & 0.591 \\
Aortic atherosclerosis & 0.608 & 0.605 & 0.620 & 0.629 & 0.639 & 0.626 & 0.638 & 0.636 & 0.615 & \underline{0.640} & \textbf{0.643} & 0.621 \\
\bottomrule
\end{tabular}}
\endgroup
\end{table*}

\section{Predicted Masks and Organ Representations}
\label{sec:mask-quality}
Supplementary Table~\ref{tab:mask-dice} reports two-dimensional Dice between
predicted masks, thresholded at 0.5, and TotalSegmentator masks on the 3,039
CT-RATE validation volumes, using the 64 slices of the downstream evaluation
and the slices on which TotalSegmentator marks the organ. For each
abnormality, we split the volumes with a nonempty reference mask into tertiles
of matched-organ Dice and averaged the 16 within-tertile AUROCs, which gave
0.852, 0.854, and 0.851 for the low, middle, and high tertiles. In 123 volumes
with large pleural effusion identified from report text, lung Dice was 0.038
[0.028, 0.049] lower than in the other volumes, and the in-organ lung pooling
weight relative to pooling with the TotalSegmentator mask was 0.978, compared
with 0.991. Each organ appears in at least one of the 64 slices in at least
96.8\% of volumes according to TotalSegmentator, and predicted masks were never
empty, so the anatomy-weighted readout needs no fallback. On the CT-RATE
validation split, the 11 organ-specific representations have a mean pairwise
linear centered kernel alignment~\cite{kornblith2019similarity} of 0.39 (range
0.14 to 0.78).

\begin{table}[t]
\centering
\caption{Two-dimensional Dice between predicted and TotalSegmentator masks on
the CT-RATE validation split. Brackets give 95\% confidence intervals from
1,000 patient-level bootstrap resamples.}
\label{tab:mask-dice}
\begin{tabular}{lc}
\toprule
Organ & Dice \\
\midrule
Lung & 0.959 [0.957, 0.960] \\
Liver & 0.956 [0.953, 0.958] \\
Spleen & 0.921 [0.917, 0.925] \\
Heart & 0.918 [0.915, 0.920] \\
Trachea & 0.909 [0.907, 0.911] \\
Aorta & 0.908 [0.906, 0.910] \\
Kidneys & 0.898 [0.893, 0.901] \\
Stomach & 0.896 [0.892, 0.901] \\
Intestine & 0.869 [0.864, 0.874] \\
Esophagus & 0.817 [0.814, 0.820] \\
Pancreas & 0.800 [0.795, 0.805] \\
\bottomrule
\end{tabular}
\end{table}

\end{document}